%% file: main.tex
\documentclass[sn-apa,iicol]{sn-jnl}% Math and Physical Sciences Reference Style
\usepackage{graphicx}%
\usepackage{multirow}%
\usepackage{amsmath,amssymb,amsfonts}%
\usepackage{amsthm}%
\usepackage{mathrsfs}%
\usepackage[title]{appendix}%
\usepackage{xcolor}%
\usepackage{textcomp}%
\usepackage{manyfoot}%
\usepackage{booktabs}%
\usepackage{algorithm}%
\usepackage{algorithmicx}%
\usepackage{algpseudocode}%
\usepackage{listings}%

\usepackage{array}%
\usepackage{color}%
\usepackage{colortbl}%
\usepackage{framed}%
\usepackage{bm}%
\usepackage{bbm}%
\usepackage{xspace}%
\usepackage{enumitem}%
\usepackage{cite}%
\usepackage{xparse}
\usepackage{url}%
\usepackage{pifont}%

\definecolor{Gray}{gray}{0.9}
\definecolor{LightCyan}{rgb}{0.88,0.95,1}

\newcommand{\mbb}[1]{\mathbf{#1}}
\newcommand{\tit}[1]{\smallbreak\noindent\textbf{#1.}}

\newcommand{\tinytit}[1]{\noindent\textbf{#1.}}

\def \ie {\emph{i.e.}}
\def \eg {\emph{e.g.}}

\newcommand{\cmark}{\ding{51}}%
\newcommand{\xmark}{\ding{55}}%

\def \ourstiny {Ours$^{\text{tiny}}$\xspace}
\def \ourssmall {Ours$^{\text{small}}$\xspace}
\def \oursbase {Ours$^{\text{base}}$\xspace}

\theoremstyle{thmstyleone}%
\theoremstyle{thmstyletwo}%

\theoremstyle{thmstylethree}%

\begin{document}

\title[Generating More Pertinent Captions]{\centering Generating More Pertinent Captions by Leveraging\\Semantics and Style on Multi-Source Datasets}

\author*[]{\fnm{Marcella} \sur{Cornia}$^\text{1}$}\email{marcella.cornia@unimore.it}
\equalcont{\small{These authors contributed equally to this work.}}

\author[]{\fnm{Lorenzo} \sur{Baraldi}$^\text{1}$}\email{lorenzo.baraldi@unimore.it}
\equalcont{\small{These authors contributed equally to this work.}}

\author[]{\fnm{Giuseppe} \sur{Fiameni}$^\text{2}$}\email{gfiameni@nvidia.com}

\author[]{\fnm{Rita} \sur{Cucchiara}$^\text{1,3}$}\email{rita.cucchiara@unimore.it}

\affil*[1]{\orgname{University of Modena and Reggio Emilia}, \orgaddress{\city{Modena}, \country{Italy}}}

\affil[2]{\orgname{NVIDIA AI Technology Centre}, \orgaddress{\city{Bologna}, \country{Italy}}}

\affil[3]{\orgname{IIT-CNR}, \orgaddress{\city{Pisa}, \country{Italy}}}

\abstract{This paper addresses the task of generating fluent descriptions by training on a non-uniform combination of data sources, containing both human-annotated and web-collected captions. Large-scale datasets with noisy image-text pairs, indeed, provide a sub-optimal source of supervision because of their low-quality descriptive style, while human-annotated datasets are cleaner but smaller in scale. To get the best of both worlds, we propose to leverage and separate semantics and descriptive style through the incorporation of a style token and keywords extracted through a retrieval component. The proposed model avoids the need of object detectors, is trained with a single objective of prompt language modeling, and can replicate the style of human-collected captions while training on sources with different input styles. Experimentally, the model shows a strong capability of recognizing real-world concepts and producing high-quality captions. Extensive experiments are performed on different image captioning datasets, including CC3M, nocaps, and the competitive COCO dataset, where our model consistently outperforms baselines and state-of-the-art approaches.}


\maketitle

\section{Introduction}
\label{sec:intro}
\input{sections/01_introduction.tex}

\section{Related Work}
\label{sec:related}
\input{sections/02_related.tex}

\section{Proposed Method}
\label{sec:method}
\input{sections/03_method.tex}

\section{Experimental Evaluation}
\label{sec:experiments}
\input{sections/04_experiments.tex}

\section{Conclusion}
\label{sec:conclusion}
\input{sections/05_conclusion.tex}

\backmatter
\bmhead{Acknowledgments}
We thank CINECA for providing computational resources. This work has been supported by the PNRR-M4C2 project (PE00000013) ``FAIR - Future Artificial Intelligence Research'' funded by the European Commission and the PRIN ``CREATIVE: CRoss-modal understanding and gEnerATIon of Visual and tExtual content'' co-funded by the Italian Ministry of University and Research (CUP B87G22000460001).

\bibliography{bibliography}

\begin{appendices}

\section{Additional Qualitative Results}
\begin{figure*}[t]
\centering
\begin{tabular}{c}
\includegraphics[width=0.99\linewidth]{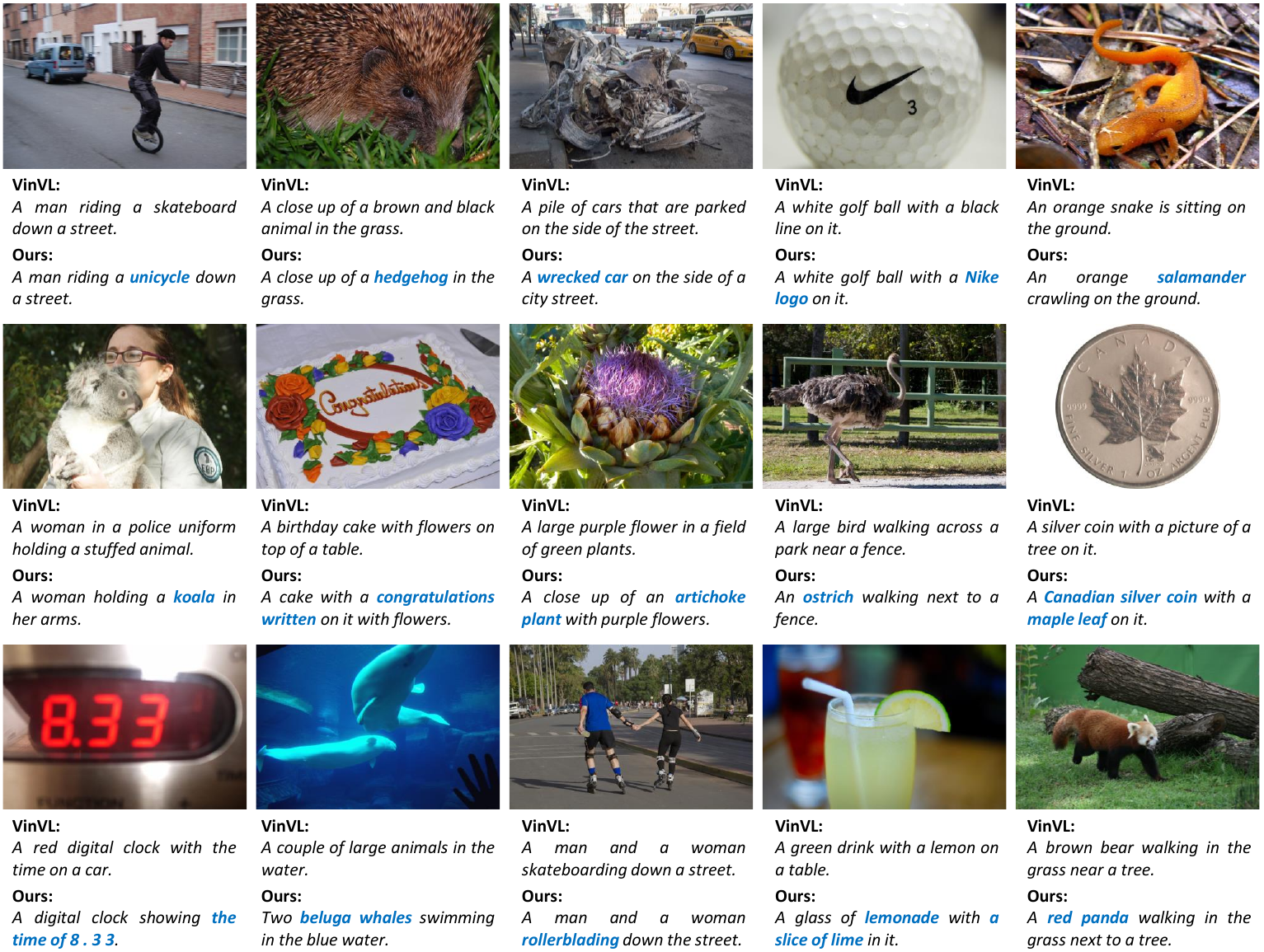} \\
\includegraphics[width=0.99\linewidth]{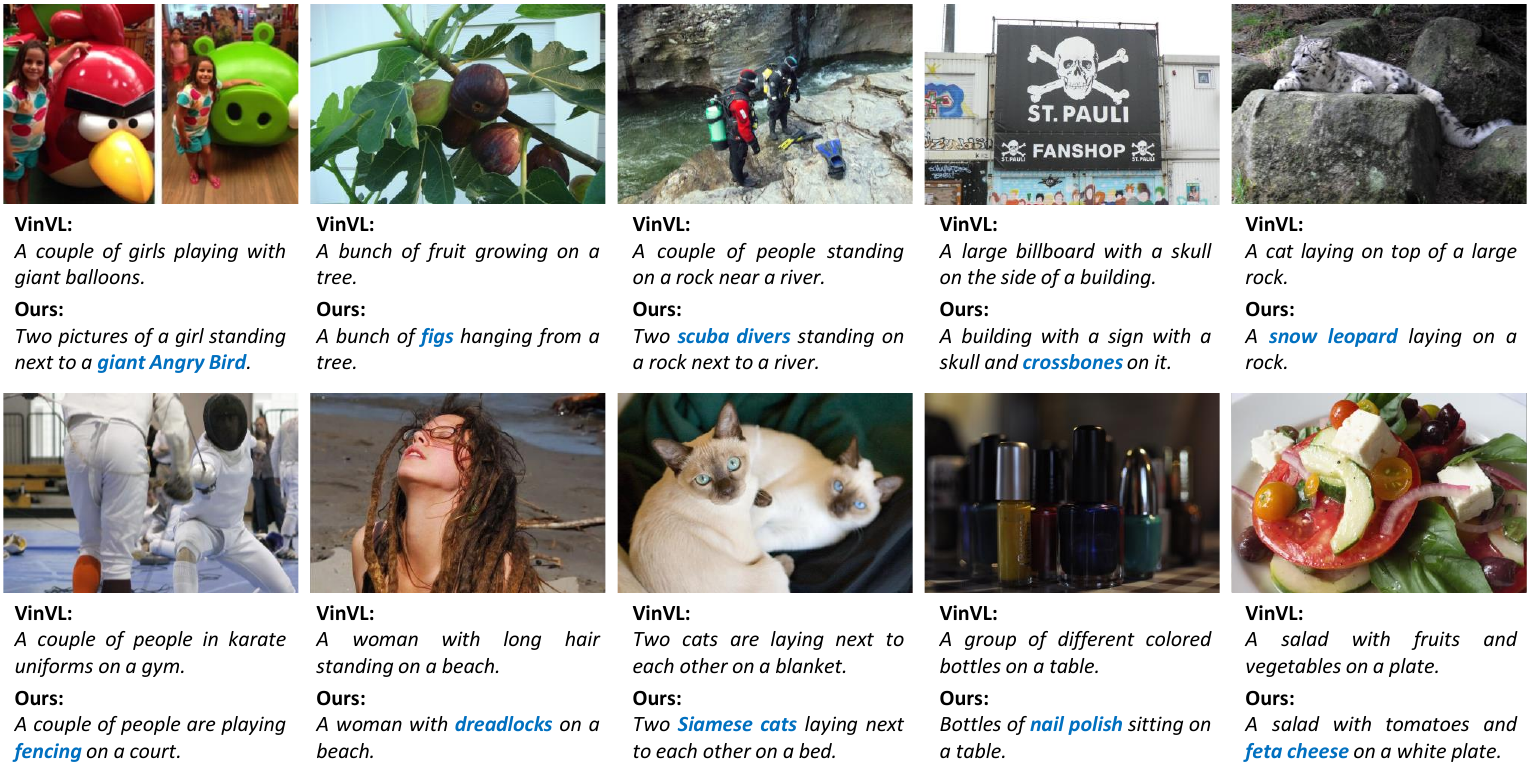} \\
\end{tabular}
\caption{Sample descriptions generated on nocaps images.}
\label{fig:nocaps_samples}
\end{figure*}

\begin{figure*}[t]
\centering
\resizebox{\linewidth}{!}{
\begin{tabular}{c}
\includegraphics[width=0.99\linewidth]{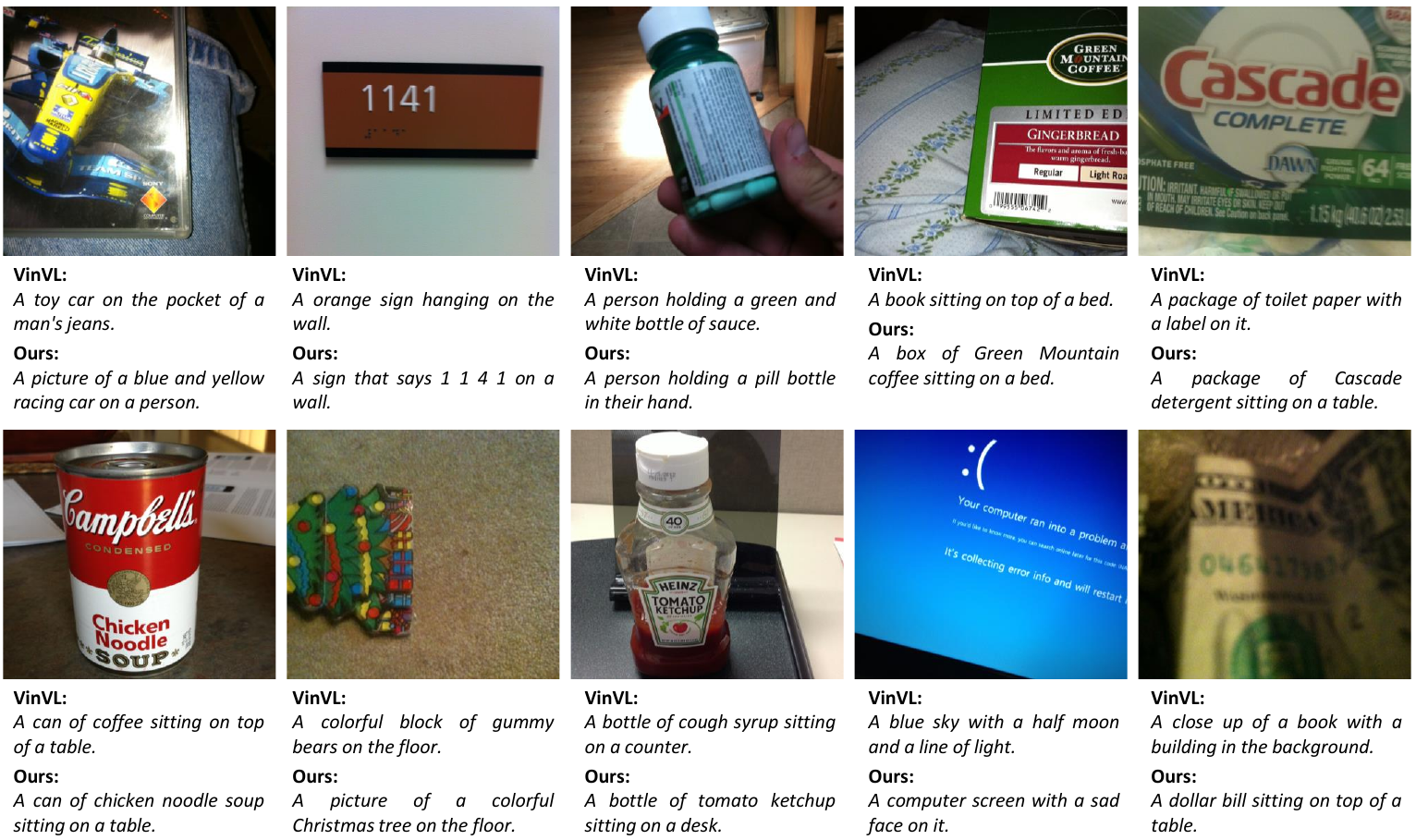} \\
\end{tabular}
}
\caption{Sample descriptions generated on images from the VizWiz dataset.}
\label{fig:vizwiz_samples}
\end{figure*}

\begin{figure*}[t]
\centering
\resizebox{\linewidth}{!}{
\begin{tabular}{c}
\includegraphics[width=0.99\linewidth]{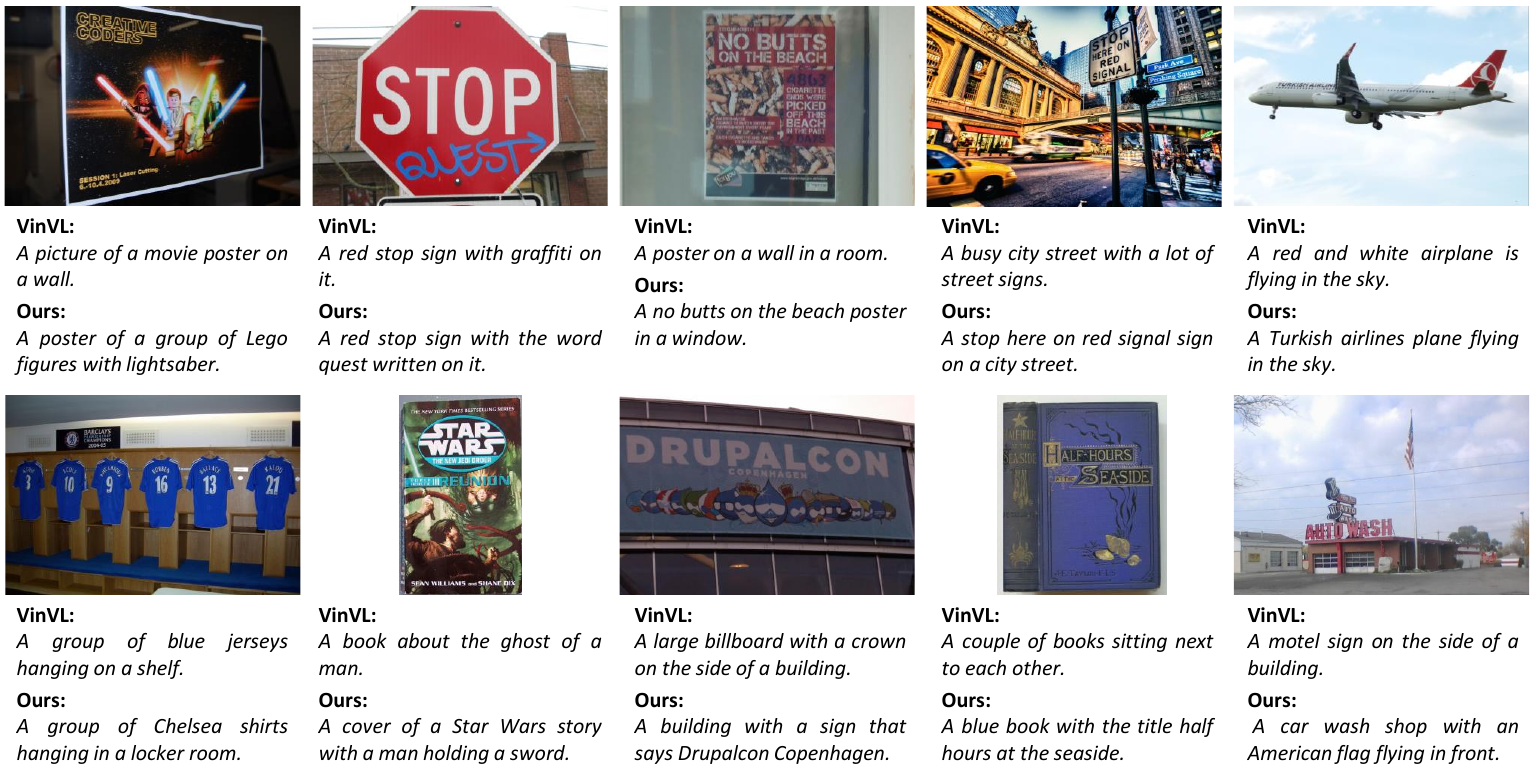} \\
\end{tabular}
}
\caption{Sample descriptions generated on images from the TextCaps dataset.}
\label{fig:textcaps_samples}
\end{figure*}

\begin{figure*}[t]
\centering
\begin{tabular}{c}
\includegraphics[width=0.99\linewidth]{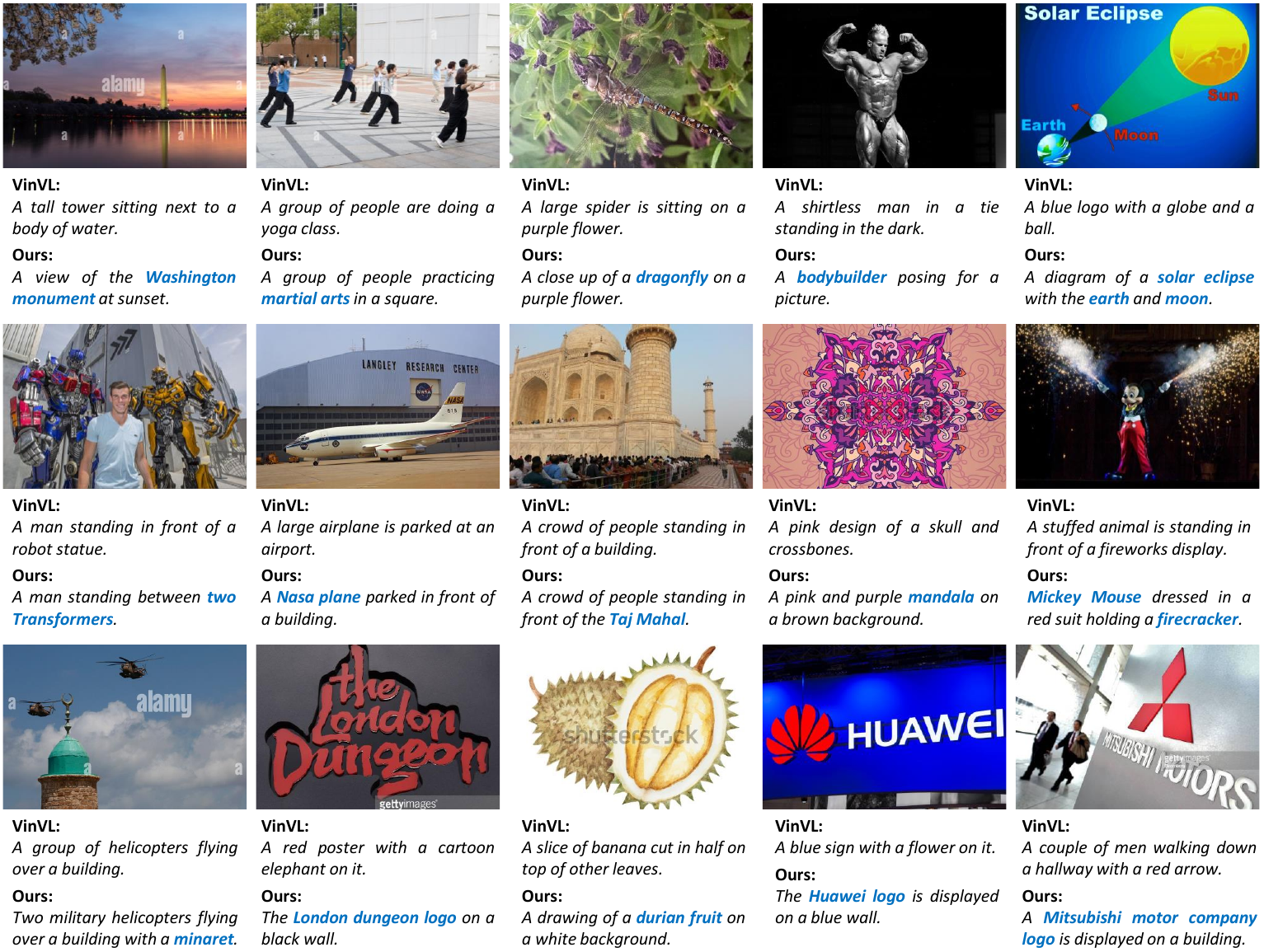} \\
\includegraphics[width=0.99\linewidth]{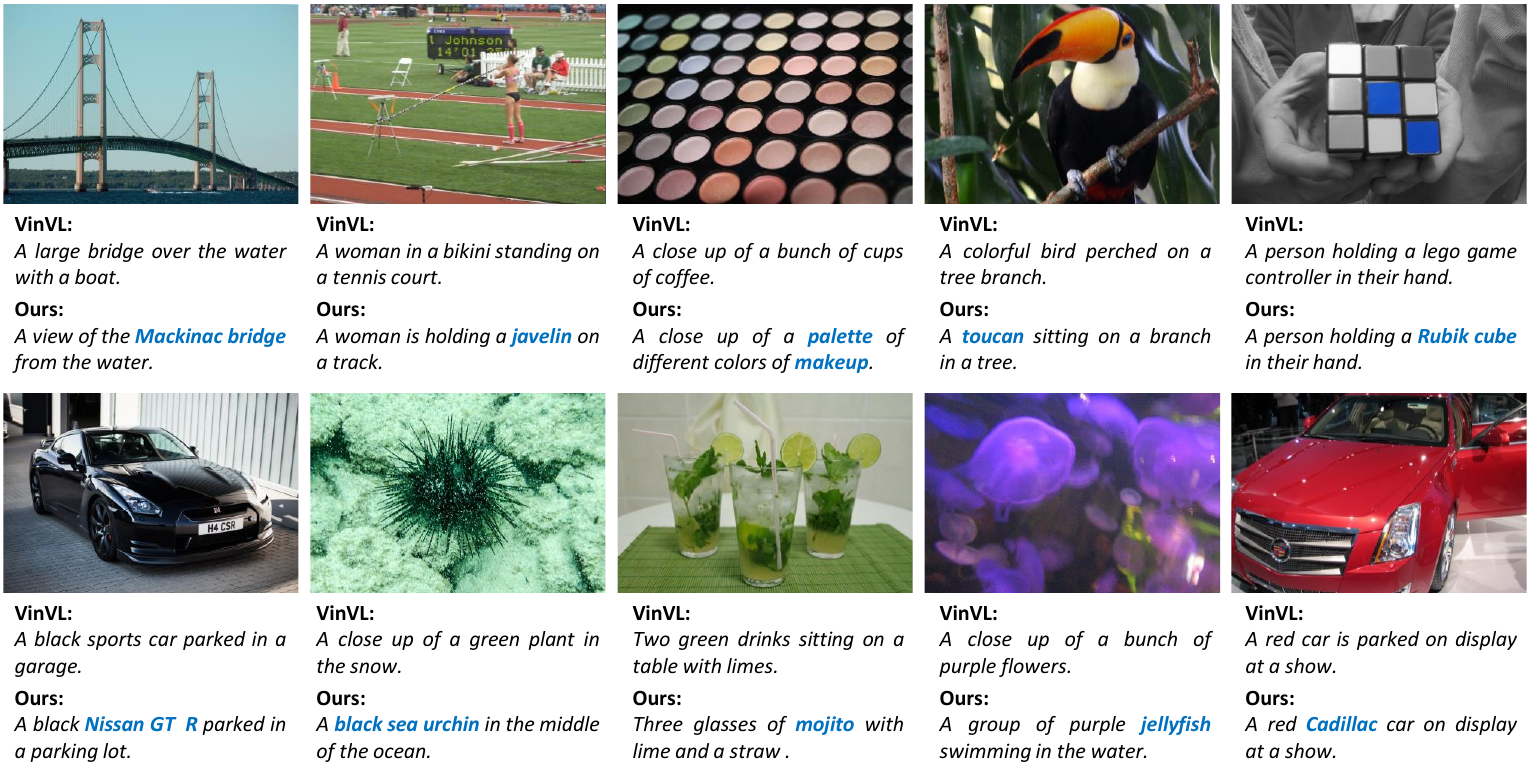} \\
\end{tabular}
\caption{Sample descriptions generated on images from the CC3M and Open Images datasets.}
\label{fig:long_tail_samples}
\end{figure*}

We report different qualitative results obtained on images from nocaps (Fig.~\ref{fig:nocaps_samples}), VizWiz (Fig.~\ref{fig:vizwiz_samples}), TextCaps (Fig.~\ref{fig:textcaps_samples}), CC3M and Open Images (Fig.~\ref{fig:long_tail_samples}). We observe how our model can describe objects, people, and scenes with a significantly increased level of detail when compared to the current state of the art and regardless of the dataset. Also, our approach qualitatively appears to be less prone to hallucination and can constantly generate fluent textual descriptions.

\end{appendices}

\end{document}

%% file: sections/01_introduction.tex
Image captioning, which aims at generating textual descriptions from visual inputs, has emerged as an attractive research problem in the last few years, as it entails modeling the connections between the visual and textual modalities~\citep{li2020oscar,zhang2021vinvl} and can be seen as a fundamental step toward machine intelligence. This has led to the development of effective strategies for feature extraction~\citep{anderson2018bottom}, cross-modal modeling~\citep{pan2020x,cornia2020meshed} and model training~\citep{rennie2017self,li2020oscar}.

\begin{figure*}[t]
\centering
\includegraphics[width=\textwidth]{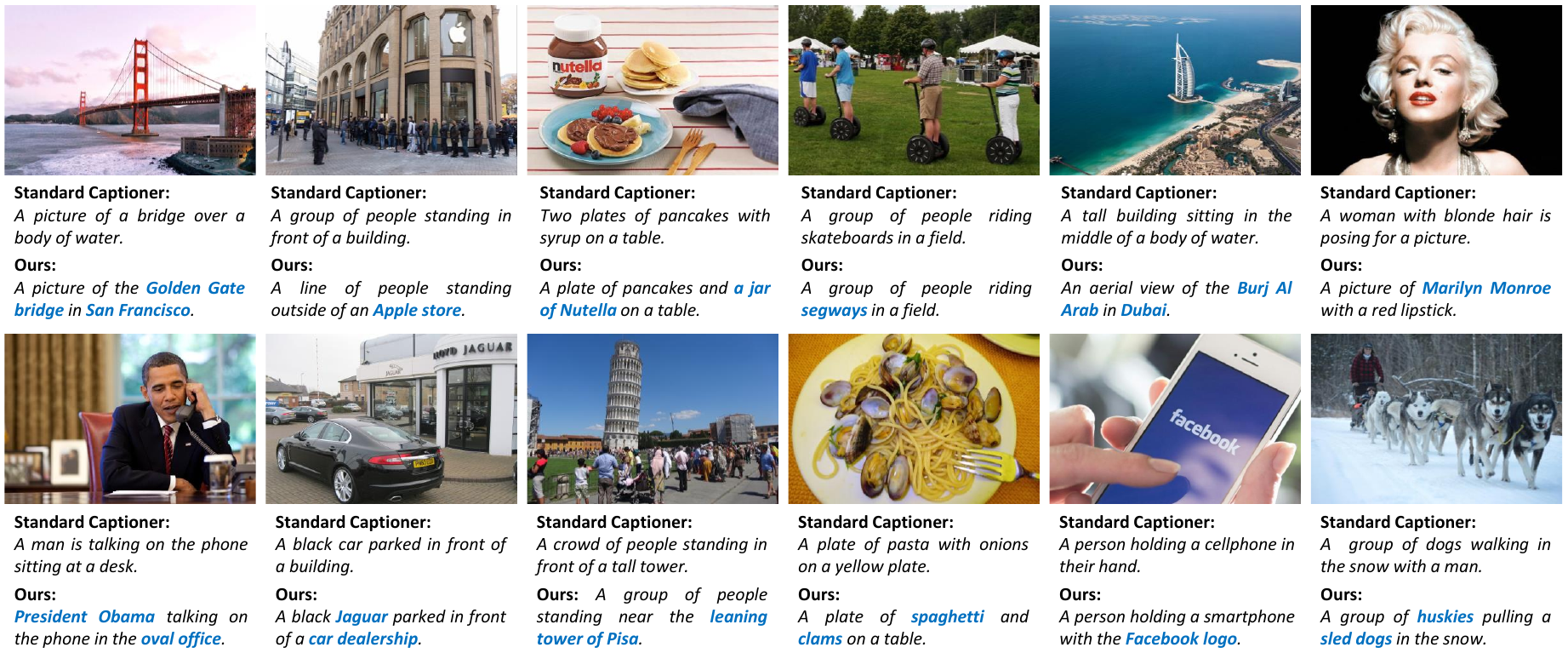}
\vspace{-0.2cm}
\caption{Sample descriptions generated by our model, in comparison with a Transformer-based captioner trained on COCO. Our approach generates high-quality captions by separating content from style.}
\label{fig:samples}
\end{figure*}

Despite these advances, researchers are still working to endow such models with the ability to describe the entire variety of real-world concepts.  The issue boils down to the limitations of popularly-used human-annotated datasets like COCO~\citep{gurari2020captioning,lin2014microsoft,young2014image} which, while being high-quality in terms of descriptive style, are limited in terms of semantic variability and size.
Recent efforts have automatically collected large-scale datasets with noisy image-text pairs from the web, partially solving the semantic scale issue at the cost of reducing the quality of the annotations~\citep{changpinyo2021conceptual,ordonez2011im2text,sharma2018conceptual,srinivasan2021wit,schuhmann2022laion}.

\begin{figure}[t]
    \centering
    \includegraphics[width=\columnwidth]{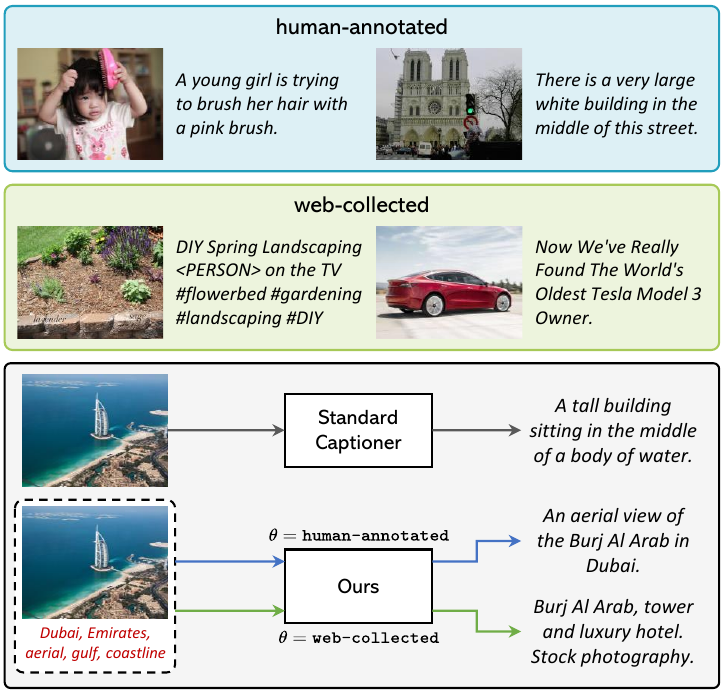}
    \vspace{-0.2cm}
    \caption{Samples of human-annotated and web-collected (image, caption) pairs and overview of our approach.}
    \label{fig:first_page}
\end{figure}

In this work we focus on generating captions that can be richer in terms of semantics and include proper names and long-tail concepts (Fig.~\ref{fig:samples}), thus being more pertinent to the input image. We do this by jointly leveraging web-collected and human-annotated sources and maintaining the style and fluency of human-annotated captions. The core idea behind our approach is that of separating semantics and descriptive style while training on non-homogeneous data sources. This is achieved through the introduction of a \emph{style token} that can condition the network both at training and generation time. During training, the token is employed to distinguish between human-annotated and web-crawled sources. At generation time, the style token can be used to generate descriptions which resemble the style of human-annotated ones, enriched by the semantics learned on web-collected datasets (Fig.~\ref{fig:first_page}).

Further, to better represent semantics, we extract textual keywords through a novel retrieval-based approach~\citep{radford2021learning}, which avoids the need of using tags or descriptions from object detectors~\citep{anderson2018bottom,zhang2021vinvl}. This also allows us to scale beyond a limited set of categories and fully represent the semantics of the image regardless of its source. The addition of the style token and of textual keywords fosters the transfer of descriptive style and semantic concepts between data sources.
 
From the point of view of the architecture, our model features an encoder-decoder structure that clearly separates the visual and textual domain, in contrast to the paradigm of employing BERT-like architectures~\citep{li2020oscar,zhou2020unified}. To represent images, we employ a multimodal feature extractor based on CLIP~\citep{dosovitskiy2021image,radford2021learning}  which can directly take raw pixels as input and avoids the need of using object detectors.
Finally, our model is trained using only a language modeling objective and does not require complex pre-training strategies.

Our model outperforms existing proposals in terms of caption quality, sometimes also surpassing models trained on significantly larger datasets~\citep{wang2021simvlm}, and shows an improved capability of generating named entities to improve the description pertinence. Experimentally, we assess the performance of the proposed approach on different image captioning datasets, including COCO~\citep{lin2014microsoft}, nocaps~\citep{agrawal2019nocaps}, VizWiz~\citep{gurari2020captioning}, and TextCaps~\citep{sidorov2020textcaps} that all contain human-annotated sentences, and CC3M~\citep{sharma2018conceptual}, WIT~\citep{srinivasan2021wit}, and a portion of LAION-400M~\citep{schuhmann2021laion} that instead are composed of web-collected data.
Overall, our work demonstrates that heterogeneous data sources can be properly exploited, together with a selective architecture, to increase the performance of image captioning systems.

\smallskip

\noindent \textbf{Contributions.} To sum up, the contributions of this paper are fourfold:
\begin{itemize}[noitemsep,topsep=5pt]
\item We propose a framework for learning on non-uniform collections of caption sources while maintaining a separation between semantic content and descriptive style. This allows the generation of fluent descriptions that resemble the quality of human-collected ones while learning from web-scale data.
\item Our approach employs a style token as a means to separate the descriptive styles of human-annotated and web-collected sources, and textual keywords extracted through a retrieval component to represent semantics.
\item In terms of architecture, our model features a fully-attentive encoder-decoder structure that jointly encodes keywords, style, and text, and is trained with a single objective of prompt language modeling.
\item We evaluate our model against carefully-designed baselines and recent approaches on COCO, nocaps, VizWiz, TextCaps, CC3M, WIT, and LAION-400M. On COCO, our approach reaches a performance of 149.6 CIDEr points.
\end{itemize}

%% file: sections/02_related.tex
\tinytit{Image Captioning} Research on image captioning has jointly focused on modeling the visual encoding pipeline, the language model, and the multi-modal connections between them~\citep{stefanini2021show}. While traditional approaches have focused on training on curated datasets, the recently emerged pre-training paradigm~\citep{chen2019uniter,lu2019vilbert,tan2019lxmert,zhou2020unified,hu2020vivo} aims at learning from weakly labeled or noisy sources. Most of the approaches have employed BERT-like~\citep{devlin2018bert} or encoder-decoder architectures in conjunction with self-supervised or sequence learning objectives. 
The OSCAR~\citep{li2020oscar} model considers triplets of object tags, detections, and captions and trains using a combination of masked token loss and contrastive loss. VinVL~\citep{zhang2021vinvl} employs the same objectives while proposing a better object detector, and trains on 8.85 million text-image pairs. SimVLM~\citep{wang2021simvlm}, instead, uses an encoder-decoder architecture and learns visual features from scratch, training on the large ALIGN dataset~\citep{jia2021scaling}. Recent models like LEMON~\citep{hu2021scaling} and BLIP~\citep{li2022blip} have further investigated the scaling properties of captioning models, also adopting custom architectures.

Recently, following the latest trends of large-scale language models~\citep{brown2020language,zhang2022opt}, several large-scale multimodal solutions have been proposed~\citep{alayrac2022flamingo,yu2022coca,li2023blip}, considerably increasing the number of parameters up to few billion (\eg~the well-known Flamingo model~\citep{alayrac2022flamingo} has 10.6B parameters) and, consequently, the computational complexity.

\tit{Visual encoders} 
In terms of visual encoding, after the emergence of global~\citep{karpathy2015deep,rennie2017self} and grid descriptors~\citep{xu2015show}, the use of object detections~\citep{anderson2018bottom,zhang2021vinvl} has become one of the most popular approaches. Indeed, it provides clean visual elements and a partial bridge between the visual and the textual domains. While several works have encoded regions through graph-based~\citep{yang2019auto} or self-attentive structures~\citep{cornia2020meshed,huang2019attention,pan2020x}, the emergence of self-attentive visual encoders~\citep{dosovitskiy2021image} and large-scale multi-modal models has enabled new strategies, ranging from training better detectors to having end-to-end visual models trained from scratch~\citep{kim2021vilt,xu2021e2e,yan2021grid,zhang2021vinvl}. 
Recently,~\citet{shen2021much} showed that features encoded by large-scale multi-modal architectures like CLIP~\citep{radford2021learning} perform at least on par with detection-based approaches. These findings have been confirmed by subsequent methods~\citep{mokady2021clipcap,barraco2022unreasonable}, also augmenting the captioning model with either knowledge distillation~\citep{barraco2022camel} or retrieval components~\citep{sarto2022retrieval,li2022comprehending}.

\begin{figure*}[t]
\centering
\includegraphics[width=\textwidth]{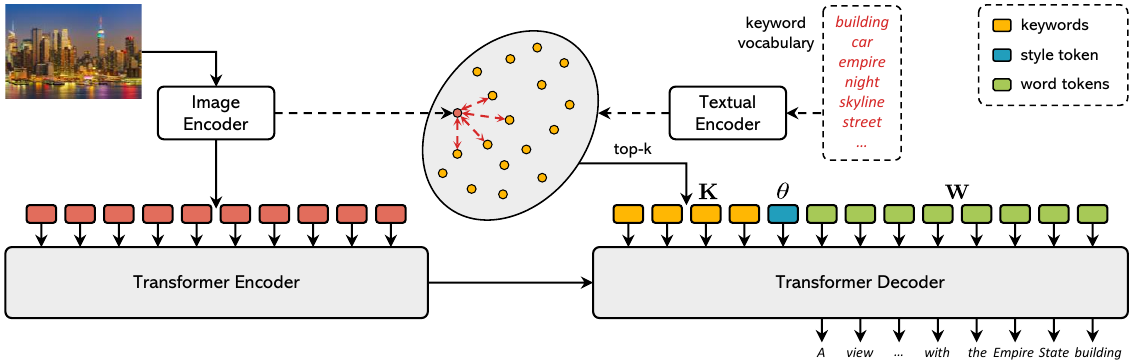}
\vspace{-0.2cm}
\caption{Illustration of the overall structure of our approach, which is composed of an encoder module, a keyword-extraction module, and a decoder module.}
\label{fig:method}
\end{figure*}%
 
\tit{Semantic content and style separation} 
A related line of research focuses on the generation of textual descriptions conditioned on a specific style or sentiment. In this context, some solutions have been proposed to train a captioning model with image-text pairs composed of images directly associated with positive and negative sentences~\citep{mathews2016senticap}, eventually constraining the captioning model to also preserve the ability to generate factual descriptions (\ie~captions without a specific style)~\citep{chen2018factual}. To reduce the dependency on paired data, other approaches proposed to leverage unpaired stylized corpus to generate more accurate stylized captions~\citep{gan2017stylenet,mathews2018semstyle,guo2019mscap,zhao2020memcap}. In particular,~\citet{guo2019mscap} employed a one-hot style indicator to condition their captioning model, which resembles our style token. Differently, the work introduced by~\citep{klein2021diverse} also considers the stylistic content of the input image represented with style-specific attributes.

While these solutions can provide a good strategy to personalize the generation of a captioning model thus potentially meeting the requirements of different users, only a few attempts have been done on the separations of semantics and style. For example, recent works on dataset collection like Conceptual Captions~\citep{changpinyo2021conceptual,sharma2018conceptual} have proposed strategies for filtering lower quality captions to ensure minimum quality levels. BLIP~\citep{li2022blip} has explicitly addressed quality assurance in web-scale training by proposing a learnable image-grounded caption filter. To our knowledge, we are the first to address the separation of semantics and style as a method for dealing with noisy web-collected captions.

%% file: sections/03_method.tex
The goal of an image captioning algorithm is that of modeling an autoregressive probability distribution over words conditioned on an input image, \ie
\begin{equation}
    p(\bm{w}_t|\bm{w}_{\tau<t}, \mbb{V}),
\end{equation}
where $\mbb{V}$ represents an input image and $\{\bm{w}_t\}_t$ is the sequence of words comprising the generated caption. In previous works, this is usually achieved by training a language model conditioned on visual features to mimic ground-truth descriptions.

The relation between images and word sequences, though, is far from being bijective, and the same semantic content can be described in different ways according to intent and descriptive style. The variance in descriptive style is a key element that differentiates human-annotated captions from noisy web-collected datasets. This is well testified in Fig.~\ref{fig:first_page}, which compares web-collected caption sources like Conceptual Captions~\citep{sharma2018conceptual} or YFCC100M~\citep{thomee2016yfcc100m} and human-annotated sources like COCO. The latter feature a grammatically correct, constant, and generic descriptive style. The former, instead, can have heterogeneous descriptive styles depending on their source and collection procedure. On average, though, they are more noisy and can comprise hashtags, comments and proper nouns.

Following this insight, we develop a model which is style-aware and can thus separate between the descriptive styles of the two aforementioned sets at generation time. Further, to enable the transfer of semantics learned from different sources, we employ textual keywords as a means to represent the content of an image regardless of its descriptive style. Formally, our approach considers a distribution
\begin{equation}
  \label{eq:optim_obj}
  p(\bm{w}_t|\bm{w}_{\tau<t}, \mbb{V}, \theta, \mbb{K}),  
\end{equation}
where $\theta$ is a parameter encoding style and $\mbb{K}$ a set of keywords encoding semantics.

\subsection{Leveraging semantics and style}

\tinytit{Extracting textual keywords through retrieval} Extracting a condensed textual representation of the visual input aims to promote an objective transfer between visual and textual features. Previous works have employed tags coming from an object detector~\citep{anderson2018bottom,zhang2021vinvl}, which however are limited in terms of the number of classes. Given the semantic breadth of web-scale datasets, it is instead crucial to scale beyond the limitation of object detection classes~\citep{krishnavisualgenome}. To this end, we cast the tagging problem as a cross-modal retrieval one, instead of a classification one. 

Given a dictionary of keywords $\mbb{Q}$, the set of predicted keywords for an input image $\mbb{V}$ is obtained by selecting the $k$ elements in $\mbb{Q}$ with the highest similarity to $\mbb{V}$, according to the matching function defined by a cross-modal retrieval model. Formally, being $\phi$ the similarity function defined by the cross-modal model, the set of keywords $\mbb{K}$ is computed as
\begin{equation}
    \mbb{K} = \underset{\bm{q} \in \mbb{Q}}{\text{argtop-}k}~ \phi(\mbb{V}, \bm{q}),
\end{equation}
where $\text{argtop-}k_{x \in \mathcal{S}} f(x)$ returns the elements in $\mathcal{S}$ which produce the $k$ largest values in $\{ f(x), x \in \mathcal{S}\}$.

The keywords dictionary $\mbb{Q}$ must be large enough to ensure sufficient coverage with respect to the semantic distribution of web-collected datasets. To this aim, we construct $\mbb{Q}$ by mining around 11.5k unigrams from COCO, Visual Genome, and the OpenWebText corpus, a public clone of OpenAI's WebText dataset~\citep{radford2019language}.
For computing cross-modal similarities, we leverage the multi-modal embedding space of CLIP~\citep{radford2021learning}, which can scale well in terms of the number of concepts. Embeddings for the keywords in $\mbb{Q}$ are obtained through the CLIP text model and can be pre-computed in advance. The embedding of an image is instead obtained through the CLIP image model and acts as a query of a $k$NN search. Although CLIP's encoder was trained on web-collected sentences, we found it to work well enough also in our case, in which it is fed with only one unigram. For efficiency reasons, we optimize this process using an index for approximate search~\citep{johnson2019billion}. 
Compared to the tags extracted from object regions~\citep{anderson2018bottom,zhang2021vinvl}, the keywords we extract do not refer to local regions of the input image, but rather correspond to the image as a whole and have an increased semantic coverage with respect to object detectors trained on standard datasets.

\tit{Style token} To aid the generation process and separate content from style, we give the model awareness of the kind of dataset to which every training caption belongs. This is done with a ``style token'', which is implemented through a learnable token embedding and which can be concatenated to the representation of the keywords. For simplicity, we employ a style token with two possible values, one for encoding human-annotated sources ($\theta = \texttt{human-annotated}$) and the second for encoding web-collected sources ($\theta = \texttt{web-collected}$). Notice that, while $\theta = \texttt{human-annotated}$ actually refers to a uniform annotation style, $\theta = \texttt{web-collected}$ can be thought as containing a collection of heterogeneous descriptive styles, as web-collected sources might have different styles (\eg, news captions and captions extracted from social media sites).

\subsection{Architecture}
Our approach represents each training image-caption pair as a quadruple $(\mbb{V}, \mbb{W}, \mbb{K}, \theta)$ of image, ground-truth caption, keywords, and style token, where $\mbb{V}$ is encoded with a set of fixed-length visual descriptors. The text input, including the caption and keywords, are tokenized lower-cased Byte Pair Encoding (BPE)~\citep{sennrich2015neural}.

For multimodal fusion, we define an encoder-decoder Transformer architecture~\citep{vaswani2017attention} where each layer of the encoder comprises multi-head self-attention (MSA) and feed-forward layers, and each layer of the decoder includes multi-head self-attention (MSA), multi-head cross-attention (MSCA), and feed-forward layers. To enable text generation, the decoder employs sequence-to-sequence attention masks in each self-attention layer. The visual descriptors $\mbb{V}=\{ \bm{v}_i \}_{i=1}^N$ are encoded via bi-directional attention in the encoder, while textual keyword tokens $\mbb{K} = \{ \bm{k}_i \}_{i=1}^M$, token embeddings of the caption $\mbb{W} = \{ \bm{w}_i \}_{i=1}^L$, and the style token $\theta$ are inputs of the decoder, where $N$, $M$, and $L$ indicates the number of visual embeddings, keywords, and caption tokens respectively. The overall network operates according to the following schema: 
\begin{align}
    \text{encoder} \quad \quad & \bm{\tilde{v}}_i = \text{MSA}(\bm{v}_i, \mbb{V}), \quad \mbb{\tilde{V}} = \{ \mbb{\tilde{v}}_i \}_{i=1}^N \nonumber \\
  \addlinespace[0.08cm]
    \text{decoder} \quad \quad & 
    \begin{aligned}
    \mbb{O}_{\bm{k}_i} &= \text{MSCA}(\bm{k}_i, \mbb{\tilde{V}}, \mbb{K}) \nonumber \\
    \mbb{O}_{\theta} &= \text{MSCA}(\theta, \mbb{\tilde{V}}, \mbb{K} \cup \theta) \nonumber \\
    \mbb{O}_{\bm{w}_i} &= \text{MSCA}(\bm{w}_i, \mbb{\tilde{V}}, \mbb{K} \cup \theta \cup \{ \bm{w}_t \}_{t=1}^i), \\
    \end{aligned}
\end{align}
where $\text{MSA}(\bm{x}, \mbb{Y})$ indicates a self-attention with $\bm{x}$ mapped to query and $\mbb{Y}$ mapped to key-values, and $\text{MSCA}(\bm{x}, \mbb{Y}, \mbb{Z})$ is a self-attention with $\bm{x}$ as query and $\mbb{Z}$ as key-values, followed by a cross-attention with $\bm{x}$ as query and $\mbb{Y}$ as key-values. $\mbb{O}$ indicates the network output and $\cup$ indicates concatenation. We omit feed-forward layers and the dependency between consecutive layers for ease of notation.

\begin{table*}[t]
\footnotesize
\begin{center}
\caption{Statistics on the training corpus.}
\label{tab:datasets}
\setlength{\tabcolsep}{.3em}
% \resizebox{\linewidth}{!}{
\begin{tabular}{lccccc}
\toprule
\textbf{Source} & \textbf{Type} & \textbf{\# Images} & \textbf{\# Words} & \textbf{\# Words in 0.1\% Tail} & \textbf{Length (mean $\pm$ std)} \\
\midrule
COCO        & Human annotated     & 112k  & 26,815    & 5,947   & 10.50 $\pm$ 2.42  \\
Flickr30k   & Human annotated    & 29k & 17,798    & 1,793   & 12.34 $\pm$ 5.21  \\
Open Images  & Generated (COCO)        & 1.7M & 7,693     & 4,050   & 10.09 $\pm$ 1.59  \\
SBU         & Flickr desc. & 875k  & 222,393   & 10,053  & 12.20 $\pm$ 6.10  \\
WIT         & Wikipedia desc. & 3.1M  & 905,095   & 281,32  & 9.21  $\pm$ 8.49  \\
CC3M        & Alt-texts         & 3.1M  & 47,422    & 21,100  & 9.58  $\pm$ 4.30  \\
CC12M       & Alt-texts         & 12.2M & 450,594   & 189,792 & 18.28 $\pm$ 13.59 \\
YFCC100M    & Alt-texts         & 14.6M & 2,384,078 & 383,942 & 26.31 $\pm$ 70.48 \\
\midrule
\textbf{Overall} & & \textbf{35.7M} & \textbf{3,180,785} & \textbf{666,519} & \textbf{19.40 $\pm$ 47.11} \\
\bottomrule
\end{tabular}
% }
\end{center}
\end{table*}

Unlike a traditional decoder, the network is only trained to predict a left-shifted version of the caption tokens $\mbb{W}$, while the sequence $\mbb{K} \cup \theta$ is treated as a prompt. Different from prompting in pre-trained language models~\citep{radford2019language,gao2020making}, this prompting strategy is explicitly employed while training the network.
Further, in contrast to previous V\&L pre-training works which adopted a bidirectional Masked Language Modeling objective that tends to be suboptimal for sequence generation, we train our network  by following a unidirectional language modeling loss based on cross-entropy, \ie
\begin{equation}
\mathcal{L} = -\mathbb{E}_{(\mbb{V}, \mbb{W}) \sim \mathcal{D}} \left( \sum_{t=1}^L \log p(\mbb{O}_{\bm{w}_t}|\mbb{V}, \mbb{K}, \theta, \bm{w}_{\tau < t}) \right),
\end{equation}
where $\mathcal{D}$ indicates the training dataset. 

In the training stage, human-annotated and web-collected image-text pairs are fed through the model, each with its corresponding style token. To allow content transfer between the two types of data sources, it is important to maintain a sufficient balance between the two kind of data sources. To this end, we randomly select samples in each mini-batch to have at least 10\% of image-caption pairs with $\theta=\texttt{human-annotated}$. In our preliminary experiments, we verified that such a percentage is enough to enable a smooth content transfer between different data sources.

\tit{Inference} Once the model is trained, predictions are conditioned on the style token $\theta$, which can be chosen according to the desired generation style (\ie~that of human-annotated captions or that of noisy web-collected ones). Given keywords and style token, at each time step $t$ the model samples a token $\bm{\hat{w}}_t$ from the output probability distribution. This is then concatenated to previously predicted tokens to form a sequence $\{ \bm{\hat{w}}_\tau \}_{\tau=1}^t$ which is employed as the input for the next iteration. Since the representation of output tokens does not depend on subsequent tokens, the past intermediate representations are kept in memory to avoid repeated computation and increase efficiency at prediction time.

\tit{Visual features}
To obtain the set of visual features $\mbb{V}$ for an image, we employ the same visual CLIP model employed for keyword retrieval~\citep{radford2021learning}. Compared to using features extracted from object detectors~\citep{hu2021scaling,zhang2021vinvl}, this strategy is beneficial both in terms of computational efficiency and feature quality. 
Specifically, we use a ViT-based visual encoder. In the original CLIP model, activations from the last layer of the encoder are discarded, except for those generated by the first query of the input sequence which are used to form a global descriptor. While global image vectors coming from CLIP have been used in concurrent captioning works~\citep{mokady2021clipcap}, we instead employ the entire grid of features coming from the layer, so to preserve spatial awareness and better feature granularity.

%% file: sections/04_experiments.tex
We conduct extensive experiments to validate the architectural choices behind our model and compare its performances with state-of-the-art solutions for image captioning.

\subsection{Training sources}
We train on a mixture of datasets with image-caption pairs, which are heterogeneous in terms of style and semantics, for a total of 35.7M images. Our mixture contains COCO~\citep{lin2014microsoft}, Flickr30k~\citep{young2014image}, SBU~\citep{ordonez2011im2text}, Conceptual Captions 3M~\citep{sharma2018conceptual} and 12M~\citep{changpinyo2021conceptual}, WIT~\citep{srinivasan2021wit}, a subset of YFCC100M~\citep{thomee2016yfcc100m}, and a subset of Open Images~\citep{kuznetsova2018open}.

The SBU dataset contains captions automatically collected from the Flickr website, while Conceptual Captions 3M and 12M have been obtained by cleaning image alt-text pairs from the web. The WIT dataset, instead, contains images extracted from Wikipedia together with alt texts. After filtering out all non-English pages, the dataset contains about 3.1M pairs. Finally, we use the subset of YFCC100M~\citep{thomee2016yfcc100m} containing image descriptions (around 14.6M pairs), and 1.7M images from Open Images~\citep{kuznetsova2018open}, automatically annotated with captions generated from OSCAR, following~\citep{zhang2021vinvl}.

In Table~\ref{tab:datasets}, we report detailed statistics on the data sources employed during training. Overall, the mixture used for training our model has three key features: (\textit{i}) differently from the datasets employed in concurrent works~\citep{li2020oscar,zhang2021vinvl}, it contains only data for the image-captioning task, thus neglecting the use of data from ancillary tasks like VQA or GQA; (\textit{ii}) it is made of publicly available data, thus allowing reproducibility, and does not employ proprietary data~\citep{jia2021scaling,wang2021simvlm}; (\textit{iii}) overall, it contains 35.7 million images and around 0.6M long-tail words (\ie~that lie in the 0.1\% of the distribution tail), making it sufficiently large and diverse to perform web-scale analyses on image captioning.

During training, we employ the style token for human-annotated sources ($\theta = \texttt{human-annotated}$) when dealing with samples from COCO, Flickr30k, and Open Images, and the one for web-collected sources ($\theta = \texttt{web-collected}$) when dealing with samples from SBU, WIT, CC3M, CC12M, and YFCC100M.

\subsection{Implementation details}
\tinytit{Architectural details}
We devise three model configurations, varying the number of decoding layers $L$, model dimensionality $d$, and number of attention heads $H$: Tiny ($L=3$, $d=384$, $H=6$, 52M params), Small ($L=6$, $d=512$, $H=8$, 87M params), and Base ($L=12$, $d=768$, $H=12$, 213M params). For all models, we employ CLIP-ViT-L/14 as visual feature and keyword extractor, three layers in the visual encoder, and five textual keywords. To represent words, we use lower-cased Byte Pair Encoding (BPE)~\citep{sennrich2015neural} with a 49,152 vocab size and linearly project them to the input dimensionality of the model $d$. We employ classic sinusoidal positional encodings~\citep{vaswani2017attention} to represent word positions. For efficiency, the maximum sequence length of the decoder is capped at 80.

\begin{figure}[t]
\centering
\includegraphics[width=\linewidth]{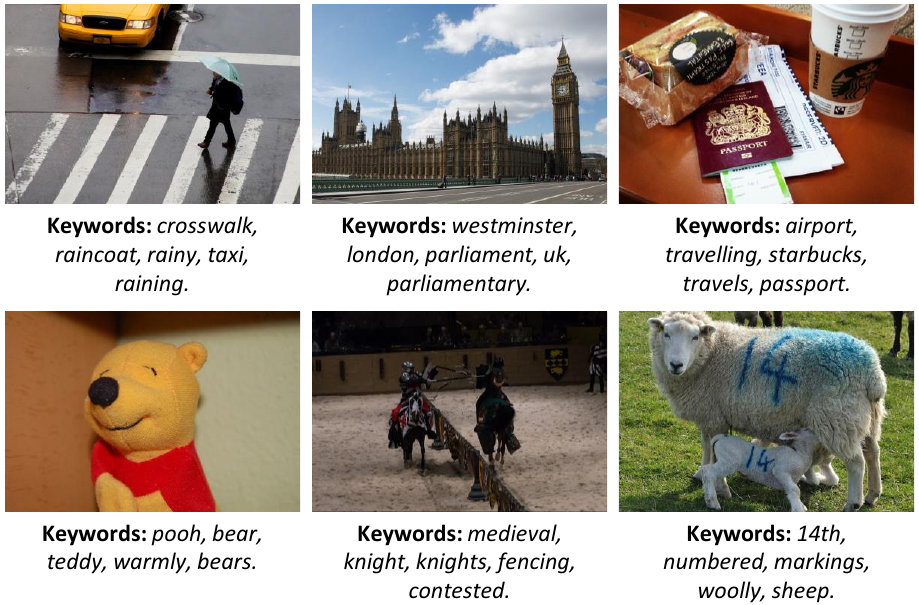}
\vspace{-0.3cm}
\caption{Sample textual keywords extracted on COCO images.}
\label{fig:keywords}
\end{figure}

Following recent literature~\citep{cornia2020meshed}, we enrich all layers of our encoder with memory slots. Specifically, we extend the set of keys and values of each self-attention operation with 40 additional learnable vectors, which are independent of the input sequence and can encode a priori information retrieved through attention. For fair comparison, this also applies to all baselines and ablation studies presented in the following.

\tinytit{Training details}
Training is performed using the LAMB optimizer~\citep{you2019large} and following the learning rate scheduling strategy of~\citep{vaswani2017attention} with a warmup equal to 6,000 iterations and multiplying the resulting learning rate by a factor of $5$. We use a minibatch size of 1,080 and employ ZeRo memory offloading~\citep{rajbhandari2020zero} and mixed-precision~\citep{micikevicius2017mixed}.
After training with cross-entropy, we also fine-tune some of our models on COCO using Reinforcement Learning. During this fine-tuning stage, we employ the SCST variant proposed in~\citep{cornia2020meshed} that sets the baseline reward equal to the mean of rewards of generated captions inside a beam. In this phase, we use the Adam optimizer~\citep{kingma2015adam}, a batch size equal to 80, and a fixed learning rate of $5\times10^{-6}$.

\tit{Keyword extraction details and visualization}
As previously mentioned, the keyword dictionary is composed by extracting around 11.5k unigrams from COCO, Visual Genome, and the OpenWebText corpus\footnote{\url{https://skylion007.github.io/OpenWebTextCorpus}}. During pre-processing, all unigrams are converted into lowercase, and proper names that identify persons are removed. Fig.~\ref{fig:keywords} reports keywords extracted from sample images of the COCO dataset. As it can be seen, they provide significant and high-level information on the global content of the image. Compared to keywords extracted from object detectors, CLIP-based keywords tend to include more long-tail concepts.

\begin{table*}[t]
\footnotesize
\begin{center}
\caption{Comparison with the state of the art on the COCO test split in a single model setting.}
\label{tab:results}
\setlength{\tabcolsep}{.25em}
\begin{tabular}{lc cccc ccccc}
\toprule
& & \multicolumn{2}{c}{\textbf{Fine-tuning}} &  & & & & & & \\ % &  & C & S \\
\cmidrule{3-4}
& & TF & SCST & \textbf{Training Images} & & B-4 & M & R & C & S \\ % &  & C & S \\
\midrule
BLIP$^\text{base}$~\citep{li2022blip} & & $\checkmark$ & - & 129M & & 39.7 & - & - & 133.3 & - \\   
BLIP$^\text{large}$~\citep{li2022blip} & & $\checkmark$ & - & 129M & & 40.4 & - & - & 136.7 & - \\   
SimVLM$^\text{base}$~\citep{wang2021simvlm} & & $\checkmark$ & - & 1.8B & & 39.0 & 32.9 & - & 134.8 & 24.0 \\ % & & 94.8 & 13.1 \\
SimVLM$^\text{large}$~\citep{wang2021simvlm} & & $\checkmark$ & - & 1.8B & & 40.3 & 33.4 & - & 142.6 & 24.7 \\ % & & 108.5 & 14.2 \\
SimVLM$^\text{huge}$~\citep{wang2021simvlm} & & $\checkmark$ & - & 1.8B & & 40.6 & \textbf{33.7} & - & 143.3 & 25.4 \\ % & & 110.3 & 14.5 \\
LEMON$^\text{base}$~\citep{hu2021scaling} & & $\checkmark$ & $\checkmark$ & 200M & & 41.6 & 31.0 & - & 142.7 & 25.1\\ 
LEMON$^\text{large}$~\citep{hu2021scaling} & & $\checkmark$ & $\checkmark$ & 200M & & 42.3 & 31.2 & - & 144.3 & 25.3 \\ 
LEMON$^\text{huge}$~\citep{hu2021scaling} & & $\checkmark$ & $\checkmark$ & 200M & & 42.6 & 31.4 & - & 145.5 & \textbf{25.5} \\ 
\rowcolor{Gray}
\textbf{\ourstiny ($\theta = \texttt{human-annotated}$)} & & - & $\checkmark$ & 35.7M & & 42.8 & 31.0 & 61.2 & 148.4 & 24.6 \\ 
\rowcolor{Gray}
\textbf{\ourssmall ($\theta = \texttt{human-annotated}$)}  & & - & $\checkmark$ & 35.7M & & 42.5 & 31.2 & 61.3 & 148.6 & 25.0 \\
\rowcolor{Gray}
\textbf{\oursbase ($\theta = \texttt{human-annotated}$)}  & & - & $\checkmark$ & 35.7M & & \textbf{42.9} & 31.4 & \textbf{61.5} & \textbf{149.6} & 25.0 \\
\midrule
OSCAR$^\text{base}$~\citep{li2020oscar} & & $\checkmark$ & $\checkmark$ & 4.1M & & 40.5 & 29.7 & - & 137.6 & 22.8 \\ 
OSCAR$^\text{large}$~\citep{li2020oscar} & & $\checkmark$ & $\checkmark$ & 4.1M & & 41.7 & 30.6 & - & 140.0 & 24.5 \\ 
VinVL$^\text{base}$~\citep{zhang2021vinvl} & & $\checkmark$ & $\checkmark$ & 5.8M & & 40.9 & 30.9 & - & 140.6 & 25.1 \\ 
VinVL$^\text{large}$~\citep{zhang2021vinvl} & & $\checkmark$ & $\checkmark$ & 5.8M & & 41.0 & 31.1 & - & 140.9 & 25.2 \\ 
\rowcolor{Gray}
\textbf{\ourstiny ($\theta = \texttt{human-annotated}$)} & & - & $\checkmark$ & 5.8M (VinVL data) & & 42.9 & 31.1 & 61.3 & 147.1 & 24.9 \\ 
\rowcolor{Gray}
\textbf{\ourssmall ($\theta = \texttt{human-annotated}$)}  & & - & $\checkmark$ & 5.8M (VinVL data) & & 42.7 & 31.3 & 61.3 & 147.5 & 25.2 \\ 
\rowcolor{Gray}
\textbf{\oursbase ($\theta = \texttt{human-annotated}$)}  & & - & $\checkmark$ & 5.8M (VinVL data) & & \textbf{43.2} & \textbf{31.4} & \textbf{61.7} & \textbf{147.8} & \textbf{25.4} \\ 
\bottomrule
\end{tabular}
\end{center}
\end{table*}

\tit{Weight initialization}
We initialize all weights by drawing inspiration from GPT-2~\citep{radford2019language}. All linear and embedding weights are initialized according to a uniform distribution and using the approach proposed by~\citep{glorot2010understanding}. Layer normalization weights are initialized to a constant value of 1. All biases are initialized to 0. We also employ the ``Special Scaled Initialization'' used in GPT-2\footnote{A reference implementation can be found in\\ \url{https://github.com/huggingface/transformers/blob/main/src/transformers/models/gpt2/modeling\_gpt2.py\#L493}} when initializing the output linear projection of each Transformer layer. Again, this also applies to all baselines.

\begin{table*}[t]
\footnotesize
\begin{center}
\caption{Results on the nocaps dataset.}
\label{tab:nocaps}
\setlength{\tabcolsep}{.3em}
\begin{tabular}{lc cccc cc c cc c cc c cc}
\toprule
& & & & & & \multicolumn{11}{c}{\textbf{Validation Set}} \\
\midrule
& & \multicolumn{2}{c}{\textbf{Fine-tuning}} & & & \multicolumn{2}{c}{\textbf{in}} & & \multicolumn{2}{c}{\textbf{near}} & & \multicolumn{2}{c}{\textbf{out}} & & \multicolumn{2}{c}{\textbf{overall}} \\
\cmidrule{3-4} \cmidrule{7-8} \cmidrule{10-11} \cmidrule{13-14} \cmidrule{16-17}
& & TF & SCST & \textbf{Training Ims} & & C & S & & C & S & & C & S & & C & S \\
\midrule
OSCAR~\citep{li2020oscar} & & $\checkmark$ & $\checkmark$ & 112k (COCO) & & 85.4 & 11.9 & & 84.0 & 11.7 & & 80.3 & 10.0 & & 83.4 & 11.4 \\
VIVO~\citep{hu2020vivo} & & $\checkmark$ & $\checkmark$ & 112k (COCO) & & 92.2 & 12.9 & & 87.8 & 12.6 & & 87.5 & 11.5 & & 88.3 & 12.4 \\
VinVL~\citep{zhang2021vinvl} & & $\checkmark$ & $\checkmark$ & 112k (COCO) & & 103.7 & 13.7 & & 95.6 & 13.4 & & 83.8 & 11.9 & & 94.3 & 13.1 \\  
\midrule
BLIP$^\text{base}$~\citep{li2022blip} & & $\checkmark$ & - & 129M & & 111.8 & 14.9 & & 108.6 & 14.8 & & 111.5 & 14.2 & & 109.6 & 14.7 \\
BLIP$^\text{large}$~\citep{li2022blip} & & $\checkmark$ & - & 129M & & 114.9 & 15.2 & & 112.1 & 14.9 & & 115.3 & 14.4 & & 113.2 & 14.8 \\
SimVLM$^\text{huge}$~\citep{wang2021simvlm} & & $\checkmark$ & - & 1.8B & & 113.7 & - & & 110.9 & - & & 115.2 & - & & 112.2 & - \\
LEMON$^\text{large}$~\citep{hu2021scaling} & & $\checkmark$ & - & 200M & & 116.9 & \textbf{15.8} & & 113.3 & 15.1 & & 111.3 & 14.0 & & 113.4 & 15.0 \\
LEMON$^\text{huge}$~\citep{hu2021scaling} & & $\checkmark$ & - & 200M & & 118.0 & 15.4 & & 116.3 & 15.1 & & 120.2 & \textbf{14.5} & & 117.3 & 15.0 \\
\rowcolor{Gray}
\textbf{\ourstiny ($\theta = \texttt{human-annotated}$)} & & - & $\checkmark$ & 35.7M & & 122.3 & 14.8 & & 115.3 & 14.6 & & 116.1 & 13.6 & & 116.5 & 14.5 \\
\rowcolor{Gray}
\textbf{\ourssmall ($\theta = \texttt{human-annotated}$)} & & - & $\checkmark$ & 35.7M & & 123.7 & 15.0 & & 118.5 & 15.0 & & 116.2 & 13.8 & & 118.8 & 14.8 \\
\rowcolor{Gray}
\textbf{\oursbase ($\theta = \texttt{human-annotated}$)} & & - & $\checkmark$ & 35.7M & & \textbf{124.8} & 15.3 & & \textbf{119.6} & \textbf{15.2} & & \textbf{120.3} & 14.4 & & \textbf{120.5} & \textbf{15.1} \\
\midrule
VinVL$^{\text{base}}$~\citep{zhang2021vinvl} & & $\checkmark$ & $\checkmark$ & 5.8M & & 112.4 & 14.7 & & 104.2 & 14.3 & & 93.1 & 12.7 & & 103.1 & 14.1 \\
VinVL$^{\text{large}}$~\citep{zhang2021vinvl} & & $\checkmark$ & $\checkmark$ & 5.8M & & 115.3 & 15.2 & & 105.6 & 14.7 & & 96.1 & 13.0 & & 105.1 & 14.4 \\
\rowcolor{Gray}
\textbf{\ourstiny ($\theta = \texttt{human-annotated}$)} & & - & $\checkmark$ & 5.8M (VinVL data) & & 121.4 & 14.9 & & 115.7 & 14.8 & & 110.6 & 13.5 & & 115.5 & 14.6 \\
\rowcolor{Gray}
\textbf{\ourssmall ($\theta = \texttt{human-annotated}$)} & & - & $\checkmark$ & 5.8M (VinVL data) & & 120.0 & 15.4 & & 117.1 & 15.2 & & 112.0 & 13.9 & & 116.5 & 15.0 \\
\rowcolor{Gray}
\textbf{\oursbase ($\theta = \texttt{human-annotated}$)} & & - & $\checkmark$ & 5.8M (VinVL data) & & \textbf{122.3} & \textbf{15.6} & & \textbf{117.7} & \textbf{15.4} & & \textbf{115.6} & \textbf{14.5} & & \textbf{118.0} & \textbf{15.2} \\
\midrule
\midrule
& & & & & & \multicolumn{11}{c}{\textbf{Test Set}} \\
\midrule
SimVLM$^\text{huge}$~\citep{wang2021simvlm} & & $\checkmark$ & - & 1.8B & & 109.0 & 14.6 & & 110.8 & 14.6 & & 109.5 & 13.9 & & 110.3 & 14.5 \\
LEMON$^\text{large}$~\citep{hu2021scaling} & & $\checkmark$ & - & 200M & & 111.2 & \textbf{15.6} & & 112.3 & 15.2 & & 105.0 & 13.6 & & 110.9 & 15.0 \\
LEMON$^\text{huge}$~\citep{hu2021scaling} & & $\checkmark$ & - & 200M & & 112.8 & 15.2 & & 115.5 & 15.1 & & 110.1 & 13.7 & & 114.3 & 14.9 \\
\rowcolor{Gray}
\textbf{\ourstiny ($\theta = \texttt{human-annotated}$)} & & - & $\checkmark$ & 35.7M & & 114.0 & 14.7 & & 115.3 & 14.7 & & 107.3 & 13.2 & & 113.7 & 14.4 \\
\rowcolor{Gray}
\textbf{\ourssmall ($\theta = \texttt{human-annotated}$)} & & - & $\checkmark$ & 35.7M & & 117.6 & 15.3 & & 117.9 & 15.0 & & 113.3 & 13.7 & & 117.1 & 14.8 \\
\rowcolor{Gray}
\textbf{\oursbase ($\theta = \texttt{human-annotated}$)} & & - & $\checkmark$ & 35.7M & & \textbf{118.8} & 15.5 & & \textbf{120.4} & \textbf{15.4} & & \textbf{114.0} & \textbf{14.1} & & \textbf{119.1} & \textbf{15.2} \\
\midrule
VinVL$^{\text{base}}$~\citep{zhang2021vinvl} & & $\checkmark$ & $\checkmark$ & 5.8M & & 104.8 & 14.8 & & 102.9 & 14.4 & & 85.8 & 12.5 & & 100.1 & 14.1 \\
VinVL$^{\text{large}}$~\citep{zhang2021vinvl} & & $\checkmark$ & $\checkmark$ & 5.8M & & 107.4 & 14.9 & & 106.2 & 14.7 & & 91.0 & 12.9 & & 103.7 & 14.4 \\
\rowcolor{Gray}
\textbf{\ourstiny ($\theta = \texttt{human-annotated}$)} & & - & $\checkmark$ & 5.8M (VinVL data) & & 115.2 & 15.2 & & 115.2 & 15.0 & & 106.3 & 13.8 & & 113.6 & 14.8 \\
\rowcolor{Gray}
\textbf{\ourssmall ($\theta = \texttt{human-annotated}$)} & & - & $\checkmark$ & 5.8M (VinVL data) & & \textbf{117.2} & \textbf{15.8} & & 115.3 & 15.1 & & 106.9 & 14.0 & & 114.0 & 15.0 \\
\rowcolor{Gray}
\textbf{\oursbase ($\theta = \texttt{human-annotated}$)} & & - & $\checkmark$ & 5.8M (VinVL data) & & 116.0 & 15.6 & & \textbf{117.4} & \textbf{15.4} & & \textbf{110.2} & \textbf{14.4} & & \textbf{115.9} & \textbf{15.2} \\
\bottomrule
\end{tabular}
\end{center}
\end{table*}

\subsection{Captioning performance}
We firstly assess the performance of our model on human-annotated datasets, comparing with recent models trained using both web-collected and human-annotated data, \ie~OSCAR~\citep{li2020oscar}, VinVL~\citep{zhang2021vinvl}, SimVLM~\citep{wang2021simvlm}, BLIP~\citep{li2022blip}, and LEMON~\citep{hu2021scaling}\footnote{The number of parameters of these models is as follows: VinVL$^\text{base}$ (135M), VinVL$^\text{large}$ (370M), LEMON$^\text{large}$ (338M), LEMON$^\text{huge}$ (675M), BLIP$^\text{base}$ (224M), BLIP$^\text{large}$ (446M), SimVLM$^\text{base}$ (86M), SimVLM$^\text{large}$ (307M), SimVLM$^\text{huge}$ (632M).}. In contrast to these approaches that employ a teacher-forcing (TF) fine-tuning on COCO when comparing on COCO and nocaps, we only employ the RL fine-tuning (SCST), which we observed being lighter in terms of forgetting impact and avoids forgetting concepts learned on web-collected data. During generation, in this setting, we set the style token $\theta$ equal to the token embedding employed for human-annotated sources.

Evaluation is reported in terms of the classical captioning metrics: BLEU~\citep{papineni2002bleu}, METEOR~\citep{banerjee2005meteor}, ROUGE~\citep{lin2004rouge}, CIDEr~\citep{vedantam2015cider}, and SPICE~\citep{spice2016}.

\tit{Performance on COCO}
The performances of our approach on COCO are reported in the upper portion of Table~\ref{tab:results}, in a single model setting. As presented, the proposed method exhibits better performances than the compared methods, without requiring a teacher-forcing fine-tuning phase, using less training data and fewer parameters. The Tiny version of the proposed approach, for instance, overcomes the performance of SimVLM, BLIP, and LEMON in their Base, Large, and Huge configurations according to the BLEU-4, ROUGE, and CIDEr metrics. Increasing model size further augment the performances, up to 149.6 CIDEr points achieved by the Base version.

As additional comparison, we also report the performance of our models when trained with a dataset size that is directly comparable to that used for OSCAR and VinVL, which also employ the SCST fine-tuning stage. In particular, we employ the same datasets used for training VinVL, excluding VQA sources. Results are reported in the bottom part of Table~\ref{tab:results}, and show increased captioning metrics with respect to the compared approaches.

\tit{Results on nocaps}
We then evaluate the capabilities of our model on the nocaps dataset~\citep{agrawal2019nocaps}, which contains out-of-domain images with respect to COCO.

\begin{table*}[t]
\footnotesize
\begin{center}
\caption{Results on the CC3M validation split.}
\label{tab:cc3m}
\setlength{\tabcolsep}{.35em}
\begin{tabular}{lc ccc ccccc}
\toprule
& & \textbf{TF Fine-tuning} & \textbf{Training Images} & & B-4 & M & R & C & S \\ 
\midrule
LEMON$^\text{base}$~\citep{hu2021scaling} & & - & 200M & & 10.1 & 11.9 & - & 108.1 & 19.8 \\ 
LEMON$^\text{base}$~\citep{hu2021scaling} & & $\checkmark$ & 200M & & 10.1 & 12.0 & - & 111.9 & 20.5 \\ 
LEMON$^\text{large}$~\citep{hu2021scaling} & & $\checkmark$ & 200M & & 10.8 & 12.3 & - & 117.4 & 21.0 \\ 
LEMON$^\text{huge}$~\citep{hu2021scaling} & & $\checkmark$ & 200M & & 13.0 & 13.9 & - & 136.8 & 23.2 \\ 
\rowcolor{Gray}
\textbf{\ourstiny ($\theta = \texttt{web-collected}$)} & & $\checkmark$ & 35.7M & & 10.6 & 13.1 & 30.0 & 121.3 & 23.0 \\ 
\rowcolor{Gray}
\textbf{\ourssmall ($\theta = \texttt{web-collected}$)}  & & $\checkmark$ & 35.7M & & 11.6 & 13.5 & 30.5 & 130.0 & 23.6 \\ 
\rowcolor{Gray}
\textbf{\oursbase ($\theta = \texttt{web-collected}$)}  & & - & 35.7M & & 9.2 & 12.1 & 27.8 & 105.7 & 20.9 \\ 
\rowcolor{Gray}
\textbf{\oursbase ($\theta = \texttt{web-collected}$)}  & & $\checkmark$ & 35.7M & & \textbf{13.2} & \textbf{14.2} & \textbf{31.4} & \textbf{144.4} & \textbf{24.7} \\ 
\bottomrule
\end{tabular}
% }
\end{center}
\vspace{-0.2cm}
\end{table*}

\begin{table*}[t]
\footnotesize
\begin{center}
\caption{Ablation study on the COCO test split and nocaps validation split.}
\label{tab:ablation}
\setlength{\tabcolsep}{.32em}
\begin{tabular}{lc ccccc c cc c ccccc c cc}
\toprule
& & \multicolumn{8}{c}{\textbf{Training Images: 35.7M}} & & \multicolumn{8}{c}{\textbf{Training Images: 5.8M (VinVL data)}} \\
\cmidrule{3-10} \cmidrule{12-19}
& & \multicolumn{5}{c}{\textbf{COCO}} & & \multicolumn{2}{c}{\textbf{\texttt{nocaps}}} & & \multicolumn{5}{c}{\textbf{COCO}} & & \multicolumn{2}{c}{\textbf{\texttt{nocaps}}} \\
\cmidrule{3-7} \cmidrule{9-10} \cmidrule{12-16} \cmidrule{18-19}
& & B-4 & M & R & C & S & & C & S & & B-4 & M & R & C & S & & C & S \\
\midrule
w/o web-collected data & & 40.6 & 30.0 & 59.9 & 139.4 & 23.9 & & 88.9 & 12.6 & & 40.6 & 30.0 & 59.9 & 139.4 & 23.9 & & 88.9 & 12.6 \\
w/o keywords and style token & & 42.5 & 30.6 & 60.8 & 145.8 & 24.2 & & 108.8 & 13.6 & & 42.4 & 30.9 & 61.0 & 145.4 & 24.8 & & 108.2 & 14.2 \\
w/o style token & & 42.5 & 30.7 & 60.9 & 146.5 & 24.3 & & 110.1 & 13.6 & & 42.5 & 31.0 & 61.1 & 146.2 & 24.8 & & 108.6 & 14.2 \\
\rowcolor{Gray}
\textbf{\ourstiny ($\theta = \texttt{human-annotated}$)} & & \textbf{42.8} & \textbf{31.0} & \textbf{61.2} & \textbf{148.4} & \textbf{24.6} & & \textbf{116.5} & \textbf{14.5} & & \textbf{42.9} & \textbf{31.1} & \textbf{61.3} & \textbf{147.1} & \textbf{24.9} & & \textbf{115.5} & \textbf{14.6} \\
\midrule
w/o web-collected data & & 40.9 & 30.4 & 60.1 & 141.5 & 24.5 & & 89.1 & 12.8 & & 40.9 & 30.4 & 60.1 & 141.5 & 24.5 & & 89.1 & 12.8 \\
w/o keywords and style token & &  42.3 & 31.0 & 60.9 & 147.5 & 24.7 & & 112.2 & 14.2 & & 42.7 & 31.1 & 61.1 & 146.2 & 25.0 & & 105.6 & 14.2 \\
w/o style token & & 42.1 & 31.0 & 61.0 & 148.1 & 24.8 & & 113.7 & 14.4 & & \textbf{43.0} & \textbf{31.3} & 61.1 & 147.2 & 25.0 & & 106.5 & 14.3 \\
\rowcolor{Gray}
\textbf{\ourssmall ($\theta = \texttt{human-annotated}$)} & & \textbf{42.5} & \textbf{31.2} & \textbf{61.3} & \textbf{148.6} & \textbf{25.0} & & \textbf{118.8} & \textbf{14.8} & & 42.7 & \textbf{31.3} & \textbf{61.3} & \textbf{147.5} & \textbf{25.2} & & \textbf{116.5} & \textbf{15.0} \\
\midrule
w/o web-collected data & & 41.4 & 30.2 & 60.2 & 142.0 & 24.1 & & 89.2 & 12.6 & & 41.4 & 30.2 & 60.2 & 142.0 & 24.1 & & 89.2 & 12.6 \\
w/o keywords and style token & & 42.6 & 31.3 & 61.2 & 147.9 & \textbf{25.1} & & 114.7 & 14.8 & & 42.6 & 31.1 & 61.2 & 146.8 & 24.8 & & 109.4 & 14.3 \\
w/o style token & &  42.5 & \textbf{31.4} & 61.2 & 149.2 & 25.0 & & 116.6 & 14.9  & & 42.7 & 31.3 & 61.5 & 147.0 & 25.2 & & 109.5 & 14.5 \\
\rowcolor{Gray}
\textbf{\oursbase ($\theta = \texttt{human-annotated}$)} & & \textbf{42.9} & \textbf{31.4} & \textbf{61.5} & \textbf{149.6} & 25.0 & & \textbf{120.5} & \textbf{15.1} & & \textbf{43.2} & \textbf{31.4} & \textbf{61.7} & \textbf{147.8} & \textbf{25.4} & & \textbf{118.0} & \textbf{15.2} \\
\bottomrule
\end{tabular}
\end{center}
\vspace{-0.1cm}
\end{table*}

Table~\ref{tab:nocaps} reports the results for both the validation and test sets, at the top and bottom part respectively. As it can be seen, the proposed method achieves higher performances with respect to previous approaches when tested on nocaps, which confirms its capability of describing out-of-domain concepts. Also in this case, our approach tends to achieve better performance using fewer parameters than competitors. Our Tiny configuration, for instance, achieves higher generation quality than SimVLM$^\text{huge}$ on both the validation and test sets, while the Small configuration is superior to LEMON$^\text{huge}$. 
\begin{figure}[t]
\centering
\includegraphics[width=\columnwidth]{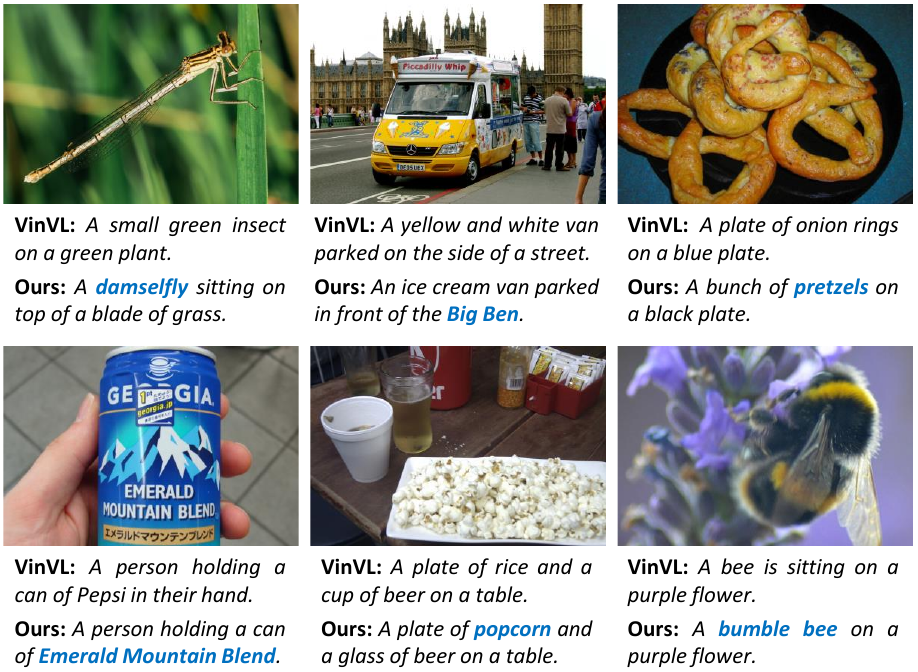}
\vspace{-0.2cm}
\caption{Comparison of captions generated by VinVL and those generated by our approach on sample images from nocaps.}
\label{fig:nocaps}
\end{figure}
It should be noted, nevertheless, that the performances on the out-of-domain portion of nocaps increase as the dimensionality of the model increases, from 116.1 to 120.3 CIDEr points when comparing the Tiny configuration to the Base configuration of the proposed approach. Overall, the Base version of our model overcomes all the reported competitors, testifying the advantage of the proposed strategies when describing out-of-domain concepts.

We also investigate the behavior of the model when trained with the same data of VinVL, and thus on a smaller dataset. Compared to VinVL, which also employs SCST fine-tuning, our model is superior according to all metrics. Scaling the dataset size from 5.8M to 35.7M images brings, however, a significant improvement: from 147.8 to 149.6 CIDEr points on COCO, and from 118.0 to 120.5 CIDEr points on nocaps. To further validate the results on nocaps, we show in Fig.~\ref{fig:nocaps} sample images and corresponding textual descriptions generated by our model (with $\theta = \texttt{human-annotated}$) in comparison to those generated by VinVL.

\tit{Results on CC3M}
We also test our approach on Conceptual Captions 3M, which contains web-collected annotations. During generation, we set the style token $\theta$ to the token embedding employed for web-collected sources, so to have an appropriate generation style. We also test our model in a zero-shot setting (\ie~without any fine-tuning), and when fine-tuning with teacher forcing on this dataset. Results are reported in Table~\ref{tab:cc3m}, in comparison with different configurations of the LEMON model. Without fine-tuning, our model achieves a better generation quality in terms of METEOR and SPICE, with slightly lower CIDEr scores. Table~\ref{tab:cc3m} also compares when running a fine-tuning stage, where our model overcomes LEMON according to all metrics, which again confirms the superiority of the proposed approach.

\begin{table*}[t]
\footnotesize
\begin{center}
\caption{Comparison with different tags sources on the COCO test split and nocaps validation split. All models are trained on the COCO dataset only.}
\label{tab:keywords}
\setlength{\tabcolsep}{.35em}
\begin{tabular}{l cc c ccccc c cc}
\toprule
& & & \multicolumn{5}{c}{\textbf{COCO}} & & \multicolumn{2}{c}{\textbf{\texttt{nocaps}}} \\
\cmidrule{4-8} \cmidrule{10-11}
& \textbf{Tags/Keywords} & & B-4 & M & R & C & S & & C & S \\
\midrule
\textbf{\ourstiny} & None & & 40.6 & 30.0 & 59.9 & 139.4 & 23.9 & & 88.9 & 12.6 \\
\textbf{\ourstiny} & Faster R-CNN & & 41.0 & 30.2 & 60.1 & 141.1 & 23.9 & & 89.8 & 12.5 \\
\textbf{\ourstiny} & ResNeXt-152 C4 & & 40.8 & 30.3 & 60.1 & 140.3 & 24.0 & & 89.9 & \textbf{12.7} \\
\rowcolor{Gray}
\textbf{\ourstiny} & Retrieval-based & & \textbf{41.1} & \textbf{30.4} & \textbf{60.2} & \textbf{141.9} & \textbf{24.1} & & \textbf{90.7} & \textbf{12.7}  \\
\midrule
\textbf{\oursbase} & None & & 41.4 & 30.2 & 60.2 & 142.0 & 24.1 & & 89.2 & 12.6 \\
\textbf{\oursbase} & Faster R-CNN & & 40.5 & 30.3 & 60.3 & 140.5 & 23.7 & & 90.2 & 13.3 \\ 
\textbf{\oursbase} & ResNeXt-152 C4 & & 40.5 & 30.2 & 60.0 & 141.7 & 23.8 & & 88.4 & 12.3 \\
\rowcolor{Gray}
\textbf{\oursbase} & Retrieval-based & & \textbf{41.6} & \textbf{30.6} & \textbf{60.7} & \textbf{143.4} & \textbf{24.2} & & \textbf{90.5} & \textbf{13.4} \\
\bottomrule
\end{tabular}
\end{center}
\vspace{-0.2cm}
\end{table*}

\begin{table*}[t]
\footnotesize
\begin{center}
\caption{Results when conditioning on different styles on the COCO test split, the CC3M and WIT validation splits, and sample images from LAION-400M. Models are trained with cross-entropy loss on the entire training corpus.}
\label{tab:indicator}
\setlength{\tabcolsep}{.35em}
\begin{tabular}{cc c ccc c ccc c ccc c ccc}
\toprule
 & & & \multicolumn{3}{c}{\textbf{COCO}} & & \multicolumn{3}{c}{\textbf{CC3M}} & & \multicolumn{3}{c}{\textbf{WIT}} & & \multicolumn{3}{c}{\textbf{LAION-400M}} \\
\cmidrule{4-6} \cmidrule{8-10} \cmidrule{12-14} \cmidrule{16-18}
& \textbf{Style token} & & B-4 & C & S & & B-4 & C & S & & B-4 & C & S & & B-4 & C & S \\
\midrule
\textbf{\oursbase} & $\theta = \texttt{web-collected}$ & & 25.7 & 86.3 & 15.5 & & \textbf{9.2} & \textbf{105.7} & \textbf{20.9} & & \textbf{2.0} & \textbf{25.9} & \textbf{6.5} & & \textbf{4.1} & \textbf{58.1} & \textbf{9.4} \\
\textbf{\oursbase} & $\theta = \texttt{human-annotated}$ & & \textbf{39.7} & \textbf{132.3} & \textbf{23.2} & & 5.0 & 58.8 & 14.6 & & 0.9 & 12.3 & 4.2 & & 2.1 & 31.2 & 6.9 \\
\bottomrule
\end{tabular}
\end{center}
\vspace{-0.1cm}
\end{table*}

\subsection{Ablation and analysis}

\tinytit{Separation of semantics and style}
In Table~\ref{tab:ablation} we investigate the role of using retrieval-based keywords and the style token. Specifically, we report the results for all three versions of our model using different amounts of image-caption pairs during training. We notice that adding keywords alone provides an improvement on both COCO and nocaps and that the benefit is especially evident on out-of-domain captioning and with large models, confirming that retrieval-based keywords help to deal with out-of-domain concepts. The combination of keywords and the style token $\theta$ for human-annotated style, instead, further boosts performances on all model configurations. For instance, when using the Tiny configuration and the entire training dataset, it increases COCO performances from 145.8 CIDEr points to 148.4. The same improvement is maintained when increasing model size, \eg~the Base configuration moves from 147.9 to 149.6 CIDEr points. Also in this case, the improvement is more significant on nocaps, where the Base configuration is improved from 114.7 to 120.5 CIDEr points, confirming that the two proposed strategies, together, help to transfer semantic concepts across descriptive styles. Similar improvements can also be noticed when training on a smaller dataset (\ie~on the same data of VinVL).

\begin{table*}[t]
\footnotesize
\begin{center}
\caption{Performances on the FlickrStyle10k and SentiCap test splits.}
\label{tab:styles}
\setlength{\tabcolsep}{.28em}
\resizebox{\linewidth}{!}{
\begin{tabular}{lc cccc c cccc c cccc c cccc}
\toprule
& & \multicolumn{4}{c}{\textbf{$\theta=\texttt{humorous}$}} & & \multicolumn{4}{c}{\textbf{$\theta=\texttt{romantic}$}} & & \multicolumn{4}{c}{\textbf{$\theta=\texttt{positive}$}} & & \multicolumn{4}{c}{\textbf{$\theta=\texttt{negative}$}} \\
\cmidrule{3-6} \cmidrule{8-11} \cmidrule{13-16} \cmidrule{18-21}
& & B-1 & B-3 & M & C & & B-1 & B-3 & M & C & & B-1 & B-3 & M & C & & B-1 & B-3 & M & C \\
\midrule
MSCap~\citep{guo2019mscap} & & 16.3 & 1.9 & 5.3 & 15.2 & & 17.0 & 2.0 & 5.4 & 10.1 & & 46.9 & 16.2 & 16.8 & 55.3 & & 45.5 & 15.4 & 16.2 & 51.6 \\
MemCap~\citep{zhao2020memcap} & & 19.9 & 4.3 & 7.4 & 19.4 & & 21.2 & 4.8 & 8.4 & 22.4 & & 50.8 & 17.1 & 16.6 & 54.4 & & 48.7 & 19.6 & 15.8 & 60.6 \\
SAN~\citep{li2021similar} & & 29.5 & 9.9 & 12.5 & 47.2 & & 30.9 & 10.9 & 13.0 & 53.3 & & 53.0 & \textbf{23.4} & 18.1 & 72.0 & & 51.2 & 20.5 & \textbf{17.6} & 67.0 \\
\midrule
\rowcolor{Gray}
\textbf{\ourstiny} & & \textbf{32.9} & \textbf{12.9} & \textbf{15.6} & \textbf{59.9} & & \textbf{34.4} & \textbf{14.3} & \textbf{15.3} & \textbf{61.6} & & \textbf{54.5} & 21.6 & \textbf{19.1} & \textbf{91.7} & & \textbf{51.4} & \textbf{22.3} & 17.5 & \textbf{83.9} \\
\bottomrule
\end{tabular}
}
\end{center}
\vspace{-0.25cm}
\end{table*}

\tit{Comparison with other keywords strategies}
We compare the proposed retrieval-based textual keywords with existing alternatives: tags extracted from Faster R-CNN trained on Visual Genome~\citep{krishnavisualgenome} and tags extracted from a ResNeXt-152 C4, trained on the same data used in VinVL. Results are reported in Table~\ref{tab:keywords} using both Tiny and Base versions, in which we train separate models with different keywords on COCO only, thus in a purely limited in-domain setting where transfer from web sources is not allowed. This choice is motivated by computational requirements, as the cost of running object detectors on large-scale data would have been intractable. As semantic transfer from web sources is not allowed, this setting is also less favorable for the proposed retrieval-based strategy. Employing the proposed tag approach, however, improves caption quality on both COCO and nocaps according to all metrics, bringing an improvement on the COCO dataset of around 0.8 and 1.4 CIDEr points respectively for the Tiny and Base versions.

\begin{figure}[t]
\centering
\includegraphics[width=\columnwidth]{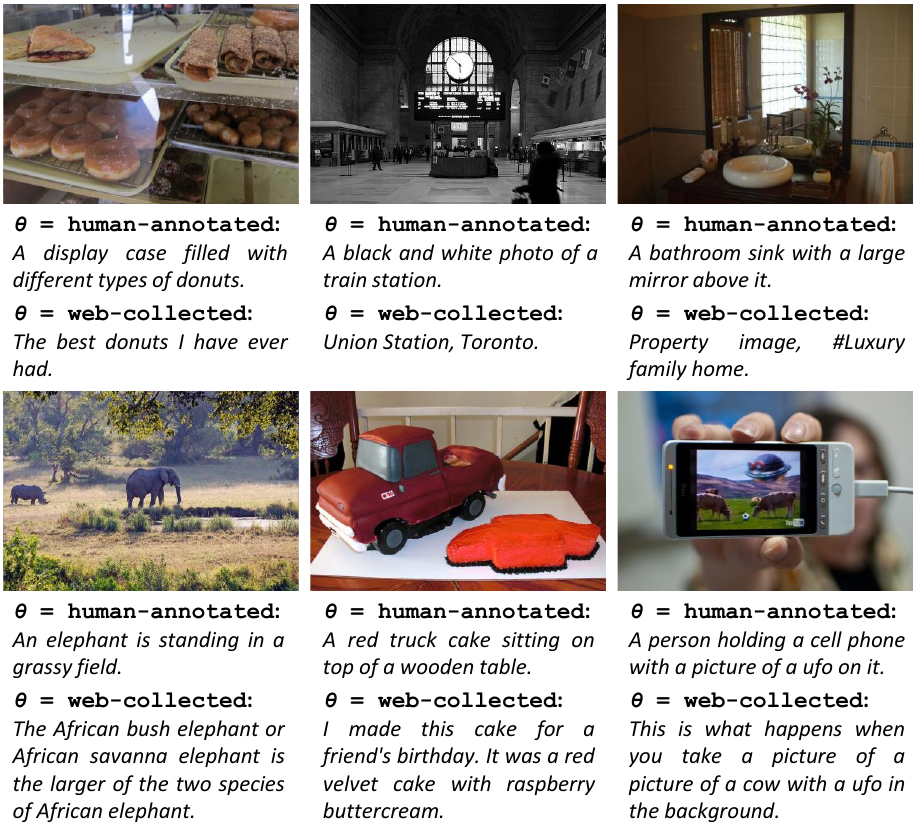}
\vspace{-0.2cm}
\caption{Sample images from the COCO dataset and captions predicted by the Base version of our model when conditioned on different styles.}
\label{fig:style_qualitatives}
\end{figure}

\begin{table*}[t]
\footnotesize
\begin{center}
\caption{User study results on sample images from the Open Images dataset. We report the percentage of times our full model (\ie~\textbf{\oursbase} with $\theta = \texttt{human-annotated}$) is preferred against a competitor. Notably, our model is always preferred more than 50\% of the times.}
\label{tab:user_study}
\setlength{\tabcolsep}{.3em}
\begin{tabular}{lc ccc}
\toprule
& & \textbf{Richness} & \textbf{Coherency} & \textbf{Fluency} \\
\midrule
vs. VinVL$^\text{large}$~\citep{zhang2021vinvl} & & 59.1 & 58.8 & 53.2 \\
\midrule
vs. \textbf{\oursbase ($\theta = \texttt{human-annotated}$)} -- w/o web-collected data & & 70.7 & 71.3 & 58.0 \\
vs. \textbf{\oursbase ($\theta = \texttt{human-annotated}$)} -- w/o keywords and style token & & 62.7 & 68.1 & 54.8 \\
\midrule
vs. \textbf{\oursbase ($\theta=\texttt{web-collected}$)} & & 53.0 & 63.6 & 67.2 \\
\bottomrule
\end{tabular}
\end{center}
\vspace{-0.1cm}
\end{table*}

\begin{table*}[t]
\footnotesize
\begin{center}
\caption{Zero-shot performances on the VizWiz test split and the TextCaps validation split.}
\label{tab:vizwiz_textcaps}
\setlength{\tabcolsep}{.26em}
\begin{tabular}{lccc ccccc c ccccc}
\toprule
& & & & \multicolumn{5}{c}{\textbf{VizWiz}} & & \multicolumn{5}{c}{\textbf{TextCaps}} \\
\cmidrule{5-9} \cmidrule{11-15}
& \textbf{Zero-Shot} & \textbf{Training Images} & & B-4 & M & R & C & S & & B-4 & M & R & C & S \\
\midrule
Up-Down~\citep{anderson2018bottom} & \cmark & 112k & & 11.3 & 12.6 & 35.8 & 18.9 & 5.8 & & 12.4 & 13.3 & 33.7 & 24.2 & 8.7 \\  
AoANet~\citep{huang2019attention} & \cmark & 112k & & 13.2 & 13.4 & 37.6 & 19.4 & 6.2 & & 18.1 & 17.7 & 41.4 & 32.3 & 11.2 \\ 
\midrule 
Up-Down~\citep{anderson2018bottom} & \xmark & 23k/22k & & 19.8 & 18.4 & 43.2 & 49.7 & 12.2 & & 20.1 & 17.8 & \textbf{42.9} & 41.9 & 11.7 \\
AoANet~\citep{huang2019attention} & \xmark & 23k/22k & & 23.2 & \textbf{20.3} & 47.1 & 60.5 & 14.0 & & 20.4 & 18.9 & \textbf{42.9} & 42.7 & 13.2 \\ 
\midrule
VinVL$^\text{base}$~\citep{zhang2021vinvl} & \cmark & 5.8M & & 16.9 & 15.8 & 41.1 & 34.7 & 9.9 & & 17.3 & 16.5 & 38.9 & 41.2 & 13.1 \\
VinVL$^\text{large}$~\citep{zhang2021vinvl} & \cmark & 5.8M & & 17.4 & 16.3 & 41.7 & 37.7 & 10.3 & & 17.5 & 16.6 & 38.9 & 41.9 & 13.1 \\
\textbf{\ourstiny ($\theta = \texttt{human-annotated}$)} & \cmark & 5.8M (VinVL data) & & 22.5 & 18.7 & 45.7 & 56.9 & 14.1 & & 20.5 & 17.9 & 41.0 & 53.0 & 14.6 \\
\textbf{\ourssmall ($\theta = \texttt{human-annotated}$)} & \cmark & 5.8M (VinVL data) & & 22.2 & 18.9 & 45.5 & 58.1 & 14.3 & & 20.7 & 18.1 & 41.1 & 54.4 & 14.7 \\
\textbf{\oursbase ($\theta = \texttt{human-annotated}$)} & \cmark & 5.8M (VinVL data) & & 22.5 & 19.2 & 45.9 & 59.6 & 14.9 & & 20.6 & 18.2 & 41.2 & 55.4 & 15.0 \\
\midrule
\rowcolor{Gray}
\textbf{\ourstiny ($\theta = \texttt{human-annotated}$)} & \cmark & 35.7M & & 23.6 & 19.3 & 46.4 & 65.6 & 14.8 & & 20.7 & 18.0 & 41.1 & 58.6 & 14.6 \\
\rowcolor{Gray}
\textbf{\ourssmall ($\theta = \texttt{human-annotated}$)} & \cmark & 35.7M & & 24.5 & 19.9 & 47.2 & 70.2 & 15.3 & & 21.9 & 18.9 & 42.3 & 66.0 & 15.4 \\
\rowcolor{Gray}
\textbf{\oursbase ($\theta = \texttt{human-annotated}$)} & \cmark & 35.7M & & \textbf{25.7} & \textbf{20.3} & \textbf{47.9} & \textbf{76.2} & \textbf{16.2} & & \textbf{23.6} & \textbf{19.4} & 42.8 & \textbf{69.9} & \textbf{15.9} \\
\bottomrule
\end{tabular} 
\end{center}
\vspace{-0.25cm}
\end{table*}

\tit{Effect of the style token}
We also assess the significance of the style token value. To this aim, we extract captions generated by our method when fed with both style token values. For this experiment, we report the results on the standard COCO test set and the CC3M and WIT validation splits. We also include the results on 30k images randomly extracted from the LAION-400M dataset~\citep{schuhmann2021laion} to validate the effectiveness of the style token on a famous web-collected dataset not employed during the training of our model. Results are reported in Table~\ref{tab:indicator}. As it can be seen, choosing the proper style token value significantly increases the quality of the generation on all considered datasets, highlighting that the model has learned to mimic both clean annotations and web-collected ones. As a complement to this experiment, we show in Fig.~\ref{fig:style_qualitatives} some qualitative results on sample images from the COCO dataset when varying the style token during inference.

To further analyze the effectiveness of our model in correctly following a given conditioning signal in the form of style tokens, we experiment with the FlickrStyle10k~\citep{gan2017stylenet} and SentiCap~\citep{mathews2016senticap} datasets which are commonly used in stylized captioning literature. While the former contains image-caption pairs with humorous and romantic styles, the latter contains images and corresponding captions with positive and negative sentiments. For fair comparison with other methods, we train the Tiny version of our model following the training protocol described in~\citep{li2021similar} and without employing web-scale data. In this case, we use a different style token for each of the four styles contained in the two datasets plus one for factual captions (\ie~captions without a specific style or sentiment). Results are shown in Table~\ref{tab:styles} comparing our model with other previous captioning approaches focused on the generation of stylized captions. Notably, our approach can also handle a larger number of styles and perform competitively compared to methods specifically designed for the stylized captioning task.

\tit{User study}
In addition to the standard quantitative metrics used in the previous analyses, we conduct a user study to fully validate the proposed model. In particular, we recruited 30 different participants and asked them to select either our full model or one of the competitors or baselines, judging in terms of (1) \textit{richness of semantics}, (2) \textit{coherency with the image}, and (3) \textit{linguistic fluency}. Participants could also state that captions were equivalent on one or more evaluation axes. In this case, 0.5 points are given to both competitors.

We perform the user study on a random subset of Open Images~\citep{kuznetsova2018open} composed of 1,000 images. As shown in Table~\ref{tab:user_study}, adding large-scale data, keywords, and the style token increases the richness of the generated captions while maintaining coherency and fluency. Moreover, the use of a cleaned style (\ie~$\theta = \texttt{human-annotated}$) also increases fluency and coherency with the input image. Even when comparing our model with the Large version of VinVL~\citep{zhang2021vinvl}, the user study results confirm the effectiveness of our approach and the improvement over the competitor especially in terms of richness of semantics and coherency with the image.

\begin{table*}[t]
\footnotesize
\begin{center}
\caption{Long-tail description performances on the validation splits of Open Images, ImageNet-21K and CC3M.}
\label{tab:longtail_results}
\setlength{\tabcolsep}{.35em}
\begin{tabular}{lcc cccc}
\toprule
& & & \multicolumn{4}{c}{\textbf{Open Images}} \\
\cmidrule{4-7} 
& \textbf{Training Images} & & Long-tail Words & Named Words & CLIP-S & PAC-S \\
\midrule
VinVL$^\text{base}$~\citep{zhang2021vinvl} & 5.8M & & 149 & 57 & 0.729 & 0.822 \\
VinVL$^\text{large}$~\citep{zhang2021vinvl} & 5.8M & & 186 & 68 & 0.736 & 0.828 \\
\rowcolor{Gray}
\textbf{\oursbase ($\theta = \texttt{human-annotated}$)} & 5.8M (VinVL data) & & 628 & 176 & \textbf{0.761} & 0.849 \\
\rowcolor{Gray}
\textbf{\oursbase ($\theta = \texttt{human-annotated}$)} & 35.7M & & \textbf{884} & \textbf{254} & 0.759 & \textbf{0.851} \\
\midrule
& & & \multicolumn{4}{c}{\textbf{ImageNet-21K}} \\
\cmidrule{4-7} 
& \textbf{Training Images} & & Long-tail Words & Named Words & CLIP-S & PAC-S \\
\midrule
VinVL$^\text{base}$~\citep{zhang2021vinvl} & 5.8M & & 149 & 64 & 0.713 & 0.813 \\
VinVL$^\text{large}$~\citep{zhang2021vinvl} & 5.8M & & 194 & 72 & 0.721 & 0.820\\
\rowcolor{Gray}
\textbf{\oursbase ($\theta = \texttt{human-annotated}$)} & 5.8M (VinVL data) & & 789 & 172 & 0.756 & 0.850 \\
\rowcolor{Gray}
\textbf{\oursbase ($\theta = \texttt{human-annotated}$)} & 35.7M & & \textbf{1152} & \textbf{261} & \textbf{0.759} & \textbf{0.851} \\
\midrule
& & & \multicolumn{4}{c}{\textbf{CC3M}} \\
\cmidrule{4-7} 
& \textbf{Training Images} & & Long-tail Words & Named Words & CLIP-S & PAC-S \\
\midrule
VinVL$^\text{base}$~\citep{zhang2021vinvl} & 5.8M & & 84 & 46 & 0.732 & 0.824 \\
VinVL$^\text{large}$~\citep{zhang2021vinvl} & 5.8M & & 95 & 45 & 0.735 & 0.829 \\
\rowcolor{Gray}
\textbf{\oursbase ($\theta = \texttt{human-annotated}$)} & 5.8M (VinVL data) & & 572 & 112 & 0.777 & \textbf{0.856} \\
\rowcolor{Gray}
\textbf{\oursbase ($\theta = \texttt{human-annotated}$)} & 35.7M & & \textbf{581} & \textbf{162} & \textbf{0.778} & \textbf{0.856} \\
\bottomrule
\end{tabular} 
\vspace{-0.1cm}
\end{center}
\end{table*}

\subsection{Zero-shot and long-tail description}
Two of the main benefits of employing web-collected data are zero-shot generalization and the description of long-tail entities. In the following, we consider the VizWiz dataset~\citep{gurari2020captioning}, which contains images originating from blind people, and TextCaps~\citep{sidorov2020textcaps}, with images containing text for zero-shot generalization. Both of them represent distinct visual and semantic distributions from the COCO ones. Further, we also investigate the capabilities of the proposed approach to deal with long-tail concepts and generate named entities.

\tit{Zero-shot performances on VizWiz and TextCaps}  Table~\ref{tab:vizwiz_textcaps} shows a comparison when using Up-Down~\citep{anderson2018bottom} and AoANet~\citep{huang2019attention} in a zero-shot manner trained on COCO and when fine-tuning them on the aforementioned datasets. We also compare with the Base and Large configurations of VinVL~\citep{zhang2021vinvl}.
As shown, the proposed approach consistently outperforms the performances of Up-Down and AoANet when evaluated in a zero-shot setting. Interestingly, it also overcomes these approaches when trained on VizWiz and TextCaps, confirming that the model is capable of properly transferring semantic concepts learned from web-collected annotations. Further, our approach also beats VinVL in both configurations by a significant margin, highlighting the appropriateness of the proposed strategies with respect to previous literature. For completeness, we also report the zero-shot performances of our approach when training on 5.8M images. Even when using less training data our approach still showcases good zero-shot prediction capabilities.

\tit{Long-tail and named entities description}
We assess the capability of our approach to name long-tail concepts and named entities. In particular, we consider the validation sets of three datasets with a large variety of concepts: Open Images V6~\citep{kuznetsova2018open} (subset with bounding boxes), ImageNet-21K~\citep{ridnik2021imagenet}, and CC3M. We count the number of unique words which do not appear in COCO at least 5 times (\ie~termed as long-tail words), and the number of named entities extracted using the spaCy NLP toolkit\footnote{\url{https://spacy.io/}}. In this setting, we evaluate caption quality using the CLIP-Score~\citep{hessel2021clipscore} and PAC-Score~\citep{sarto2023positive} metrics, that both are based on CLIP features coming from a ViT-B/32 model, do not require ground-truth captions, and have a high correlation with human judgments.
Table~\ref{tab:longtail_results} shows the results of both versions of our Base model trained on 5.8M and 35.7M images, in comparison with VinVL. Our approach is capable of naming significantly more words that are outside of COCO and also consistently generates more named entities than VinVL. This improvement is further confirmed by the results in terms of CLIP-S and PAC-S which are significantly better than those obtained by VinVL. Reducing the amount of web-collected sources impacts performances, with a significant reduction in the number of long-tail words and named entities produced during generation. We notice, though, that even when using the same dataset size our approach beats VinVL according to all evaluation metrics in both its Base and Large configurations, further confirming the appropriateness of the proposed approach, and of avoiding teacher-forcing fine-tuning.

\subsection{Qualitative results}
\begin{figure}[t]
\centering
\includegraphics[width=\columnwidth]{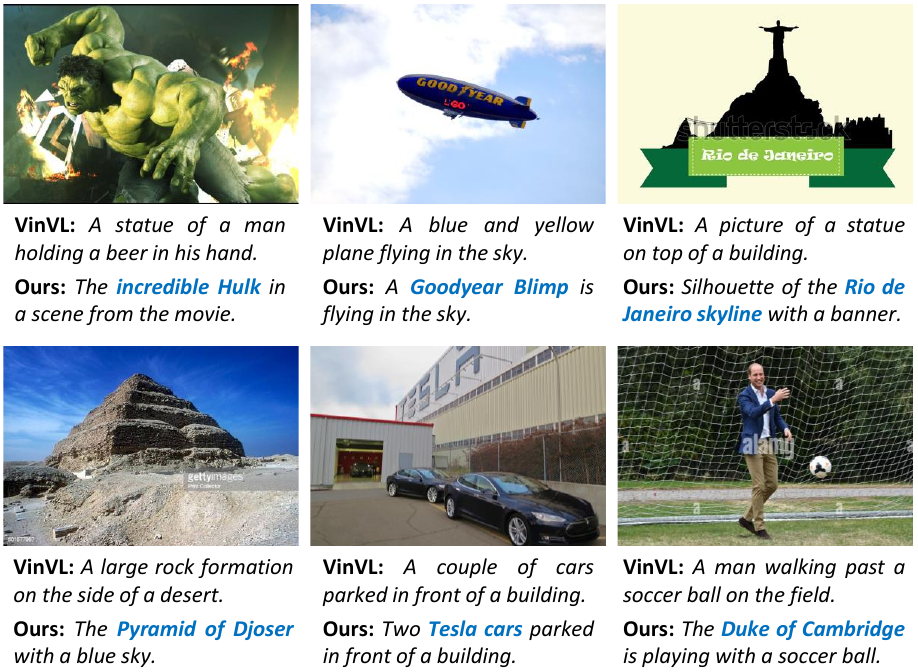}
\vspace{-0.3cm}
\caption{Comparison of captions generated by VinVL and those generated by our approach on sample images from CC3M and Open Images.}
\label{fig:qualitatives}
\vspace{-0.2cm}
\end{figure}

Finally, we showcase the capabilities of our approach of generating pertinent captions with named entities (\ie~recognizable objects, places, and people) through some qualitative examples. In Fig.~\ref{fig:samples}, on the second page, we compare with the baseline Tiny model trained without web-scale data and keywords (reported in Table~\ref{tab:ablation}). We observe that our approach correctly recognizes and names famous people, places, and trademarks like the \textit{Burj Al
Arab}, \textit{Marilyn Monroe}, or the \textit{Facebook logo}. 
The same can be observed in Fig.~\ref{fig:nocaps} and~\ref{fig:qualitatives}, where we compare with VinVL$^\text{large}$~\citep{zhang2021vinvl} on images taken from nocaps, OpenImages, and CC3M. Again, our model can recognize and name long-tail concepts better than previous approaches, also recognizing famous people and places as for example the \textit{Duke of Cambridge} and the \textit{Pyramid of Djoser}. Additional qualitative results are reported in the Appendix.

%% file: sections/05_conclusion.tex
We proposed an approach for captioning images with fluent and pertinent descriptions while training on non-uniform data sources. Our approach relies on the separation of semantics and style and the usage of retrieval-based textual keywords, and allows to learn from noisy web-collected sources while maintaining a fluent descriptive style. Experimentally, it achieves state-of-the-art results on different datasets, including COCO, CC3M and nocaps, demonstrating its effectiveness in both in-domain and out-of-domain image captioning.